# A Systematic Evaluation of Current Architectures in Wind Power Forecasting

**VINÍCIUS BORTOLINI[1,2], GILSON ADAMCZUK OLIVEIRA[1], ÉRICK OLIVEIRA RODRIGUES[1], AND MATHEUS HENRIQUE DAL MOLIN RIBEIRO[1]**

[1]Industrial and Systems Engineering Graduate Program (PPGEPS), Federal University of Technology Paraná (UTFPR), Pato Branco, Paraná 85503-390, Brazil
[2]Institute Federal of Paraná (IFPR), União da Vitória, Paraná 84603-580, Brazil

Corresponding author: Vinícius Bortolini (viniciusbortolini@alunos.utfpr.edu.br)

This work was supported in part by National Council for Scientific and Technological Development (CNPq) under Grant 314022/2023-6, and in part by the Araucária Foundation under Grant PRD2023361000550.

**ABSTRACT** Interval wind speed forecasting is essential for the efficient integration of wind energy into power systems, as it accounts for the inherent uncertainty of wind resources. This study presents a systematic literature review focused on hybrid approaches to interval forecasting of wind generation, exploring the combination of deep learning, modal decomposition, and statistical methods. To guide the paper selection, Latent Dirichlet Allocation (LDA) was applied for topic modeling, enabling the identification of patterns and research trends. The findings emphasize that integrating hybrid models with decomposition techniques—such as Variational Mode Decomposition (VMD) and Ensemble Empirical Mode Decomposition (EEMD)—enhances forecast accuracy and reliability by narrowing prediction intervals without compromising coverage. Regarding interval construction, most studies adopt a dual-model strategy, independently forecasting the lower and upper bounds. Input data are commonly decomposed using techniques like EMD, EEMD, or VMD, which extract frequency-based components. These components serve as inputs to models such as LSTM or ELM, trained separately for each bound. This approach allows for targeted modeling of uncertainty, improving flexibility and precision. Interval quality is typically evaluated through metrics that balance coverage and interval width. The review also highlights challenges, including the lack of standardized evaluation metrics, computational complexity, and limited real-world validation. Overall, the study reinforces the value of interval forecasting for wind energy operations and offers insights for advancing model robustness and decision-making.

## I. INTRODUCTION

Wind energy is a major renewable source, widely used for electricity generation worldwide, including Brazil [1]. However, forecasting wind power is challenging due to its variability and intermittency [2]. Accurate estimates are essential for grid stability, operational planning, and balancing supply and demand, making forecasting methods indispensable for efficient management [3].

According to Yao et al. [4], forecasting has been widely studied using approaches such as time series models and machine learning (ML) techniques. These methods capture historical patterns and project them into the future, enabling more precise grid operation and planning.

Forecasts can be short-, medium- or long-term; short-term forecasts are critical for daily wind farm optimization. Methods include point forecasts, which give a single value, and interval forecasts, which provide a range and incorporate model uncertainty [5].

ML predictive models can significantly improve accuracy [3]. For example, Chen et al. [3] applied Gaussian Process Regression (GPR) to address nonlinear and uncertain wind behavior, outperforming traditional statistical models by integrating meteorological variables such as wind speed and direction with uncertainty estimation. ML has also been applied in healthcare [6], economics [7], and energy demand forecasting [4], learning from large datasets and surpassing traditional point forecasting methods [3], [7].

In transmission line ampacity forecasting, Dupin et al. [8] evaluated four models—Quantile Linear Regression (QLR), Mixture Density Neural Network (MDNN), Kernel Density Estimator (KDE), and Quantile Regression Forest (QRF)—using two years of weather data for a 110 kV line in Northern Ireland. QRF reduced mean absolute percentage error (MAPE) by 28% compared to persistence for 24-hour forecasts and achieved the best probabilistic performance. Using QRF with a fixed quantile strategy increased average ampacity by up to 35% over Static Line Rating (SLR) while maintaining acceptable risk.

For voltage stability, short-term prediction combining Graph Convolutional Networks (GCN) with Long Short-Term Memory (LSTM) improves real-time monitoring [9]. In cybersecurity, ML detects and mitigates data integrity attacks in smart grids, even with concept drift [10]. Load forecasting benefits from Recurrent Extreme Learning Machines (RELM) [11], and hybrid CNN–LSTM architectures aid fault detection and localization in transmission lines [12]. These applications highlight the versatility and effectiveness of ML in power systems.

Given the large amount of studies on wind forecasting with ML and time series, extracting and analyzing relevant data is complex. An effective approach is to first identify topics of interest and then analyze related documents [13]. Topic modeling, such as Latent Dirichlet Allocation (LDA), supports this by identifying relevant topics in large text corpora and improving replicability [14]. In this work, LDA is used to identify topics related to wind energy forecasting with ML.

In recent years, several reviews have aimed to consolidate knowledge on wind power forecasting, each addressing different aspects of the field. Xie et al. [15] provided one of the most comprehensive taxonomies of deterministic and probabilistic forecasting methods, classifying models into physical, statistical, and hybrid categories and summarizing evaluation metrics such as Root Mean Square Error (RMSE), Mean Absolute Error (MAE), Prediction Interval Coverage Probability (PICP), and Coverage Width Criterion (CWC). However, their work primarily focused on establishing a broad conceptual framework for forecasting methodologies and did not examine in depth the mechanisms by which uncertainty is quantified in interval forecasting. Alongside that, exponential smothing average can be used to obtain control limits for forecasting [16].

Maldonado-Correa et al. [17] conducted a systematic literature review with a practical orientation, centered on short-term forecasting for a specific high-altitude wind farm (the Villonaco Wind Power Plant). Their study explored the use of artificial intelligence tools—particularly artificial neural networks—and emphasized operational variables from SCADA systems and meteorological inputs such as wind speed, temperature, and pressure. Although this work contributed valuable insights for site-specific forecasting, it did not address probabilistic or interval-based approaches, nor the hybrid decomposition methods that have become increasingly relevant in recent research.

More recently, Tsai et al. [18] reviewed modern artificial intelligence and deep learning methods for ultra-short- and short-term wind power prediction. Their analysis emphasized the evolution of hybrid deep learning models and their performance across different time resolutions, highlighting trends in computational complexity and accuracy. Nevertheless, this review focused mainly on model architectures and algorithmic efficiency, without providing an extensive discussion on interval or uncertainty modeling.

To clarify the conceptual scope of this review, hybrid models are herein defined as integrated forecasting frameworks that combine at least two complementary methodological components — typically a signal decomposition stage (e.g., VMD, CEEMDAN, or EMD) and a predictive component based on statistical or deep learning techniques (e.g., ARIMA, LSTM, or CNN-BiLSTM). The decomposition process separates the original time series into subcomponents with different frequency characteristics, reducing noise and enhancing feature extraction, while the predictive layer captures complex nonlinear dependencies.

In contrast to these previous studies, the present work provides a comprehensive and systematic review focused on hybrid interval forecasting approaches for wind energy. The main contribution lies in integrating interval prediction, hybrid modeling, and signal decomposition techniques (e.g., Variational Mode Decomposition (VMD), Complete Ensemble Empirical Mode Decomposition with Adaptive Noise (CEEMDAN), and Empirical Mode Decomposition (EMD)) within a unified analytical framework. By employing LDA for topic modeling, the study introduces a data-driven approach to identify prevailing research themes and emerging gaps in the field.

Moreover, this review conducts a detailed analysis of how different authors construct the prediction intervals (PIs) themselves, examining methodological formulations, the strategies used to define upper and lower bounds, and the evaluation metrics applied to assess interval quality (such as PICP and CWC). This type of in-depth methodological examination has not been thoroughly addressed in earlier reviews, which typically focused on general classifications of models or overall performance comparisons.

Consequently, this study offers a deeper and more structured understanding of the methodological foundations behind interval forecasting, complementing and extending the scope of previous literature on deterministic, probabilistic, and deep learning-based forecasting.

The remainder of the article is structured as follows: Section II describes the methodology; Section III reports the main findings; Section IV discusses results; and Section V presents conclusions and future directions.

## II. METHODOLOGY

The Systematic Literature Mapping comprised defining the research question and search strings, retrieving relevant studies, analyzing documents and keywords, and extracting data. The Orange[1] tool was employed for data preprocessing, text mining, and topic modeling with LDA.

The research question was: "How can wind energy forecasting models based on interval forecasting be applied to optimize the generation and management of energy resources?". From this, the keywords Wind Power, Time Series, Forecasting and Interval Forecasting were combined with Wind Energy, yielding the search term: ("Wind Energy" OR "Wind Power") AND ("Time Series") AND ("Forecasting" OR "Interval Forecasting"). Applied to the Scopus[2] database (2015–Oct. 2024), this resulted in 1443 documents (articles and conference papers). Only English-language studies were considered, given their predominance.

The word-co-occurrence network, subsequently generated using VOSviewer[3] from the 1,443 documents retrieved in Scopus, reveals the main thematic relationships in the field of wind power forecasting, as shown in Figure 1. The central cluster is dominated by terms such as "wind power forecasting," "machine learning," "time series," and "deep learning," highlighting the strong focus on data-driven and hybrid approaches. Peripheral clusters include terms like "VMD," "hybrid model," "uncertainty," and "interval forecasting," indicating growing interest in signal decomposition and uncertainty quantification. Overall, the map confirms that current research trends converge toward integrating deep learning and decomposition techniques to enhance the reliability and accuracy of wind energy forecasts.

A word cloud was generated from abstracts (Figure 2), supporting validation of selected terms and suggesting additional keywords for refining the query.

Preprocessing was carried out in Orange, following four steps:

1) Transformation: conversion to lowercase and removal of accents;
2) Tokenization: division into tokens;
3) Normalization: lemmatization of words;
4) Filtering: removal of stopwords, numbers and punctuation.

This ensured data uniformity and reduced noise, essential for reliable modeling.

### A. TOPIC MODELING

Topic modeling was conducted with LDA [13], [19], which groups documents by semantic similarity. It highlighted key areas such as "hybrid models," "modal decomposition," and "neural networks," guiding the selection of relevant studies. Several topic numbers (K) were tested and evaluated using coherence and perplexity [19], [20]. The best trade-off was K=3 (coherence 0.3691; perplexity 309.109), outperforming K=4 (0.3455; 299.58) and K=7 (0.3432; 280.83). Thus, K=3 was adopted, offering cohesive topics with competitive generalization (Figure 3).

Keywords extracted from each topic were:

1) speed, prediction, lstm, multi, decomposition, neural, learning, term, network, interval;

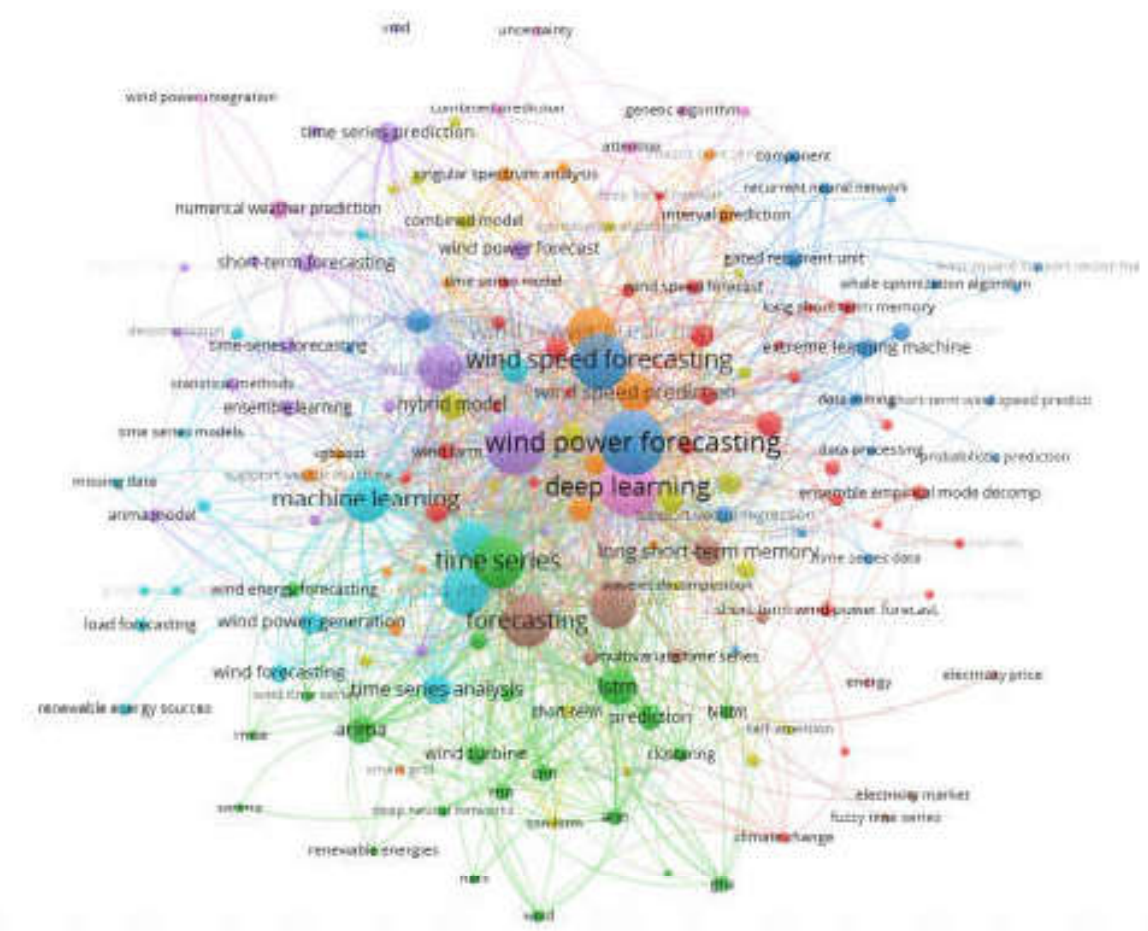

**FIGURE 1.** Keyword co-occurrence network of the 1,443 documents retrieved from scopus (Generated Using VOSviewer).

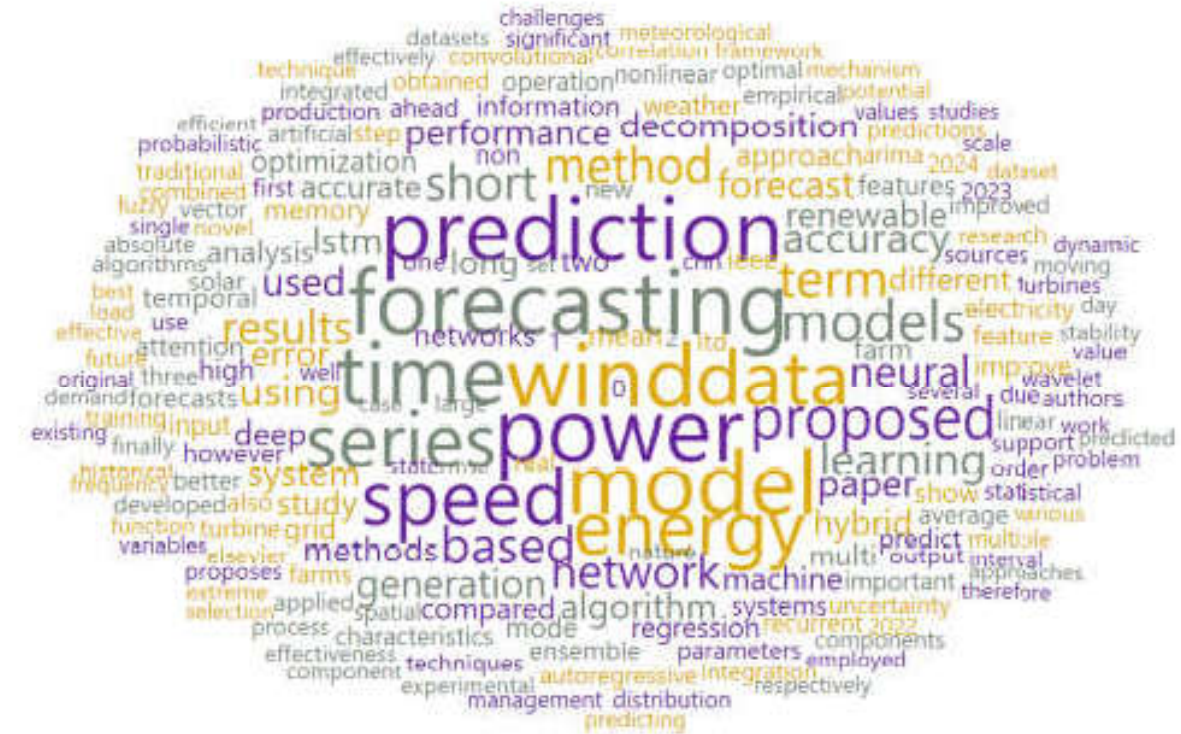

**FIGURE 2.** Word cloud of the most frequent terms in the selected abstracts.

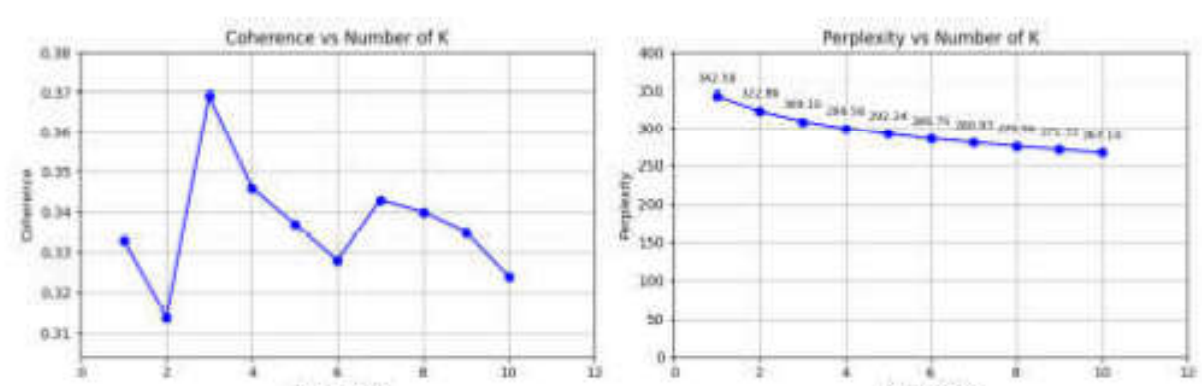

**FIGURE 3.** Analysis of the coherence and perplexity factor in topic modeling.

[1] https://orangedatamining.com/
[2] https://www.scopus.com
[3] https://www.vosviewer.com/

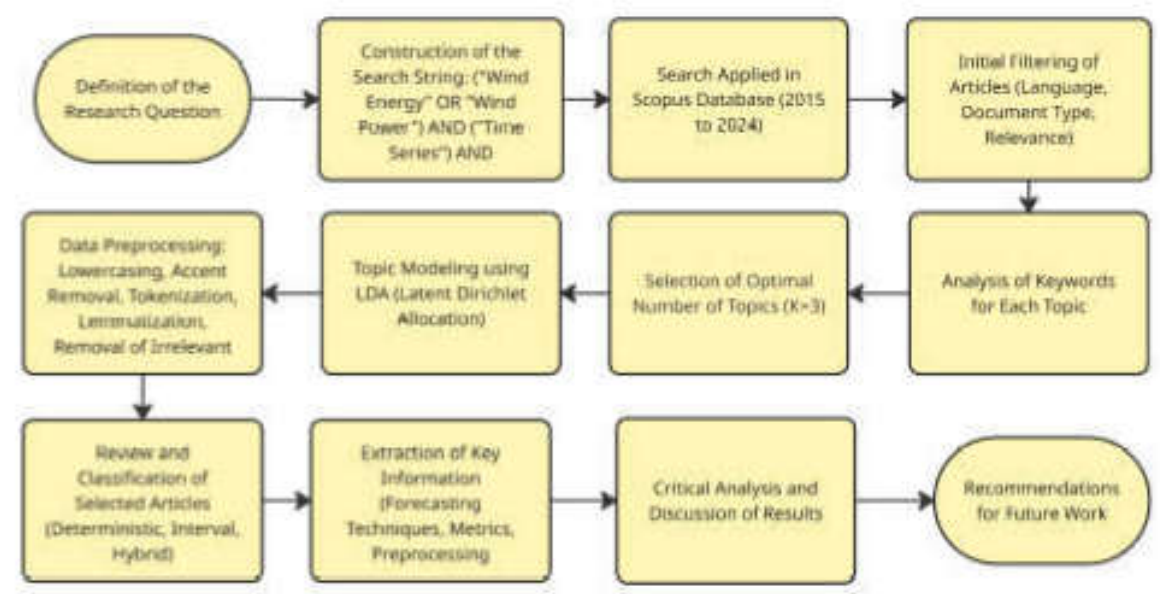

**FIGURE 4.** Diagram representing the methodological process applied in this work.

2) decomposition, speed, prediction, mode, vector, algorithm, hybrid, energy, machine, lstm;
3) speed, energy, arima, electricity, learning, generation, hybrid, system, forecast, forecasting.

The term LSTM was removed from the first two to broaden scope. Table 1 presents the labeled topics with keywords, descriptions, search terms and number of papers.

Applying these search terms in Scopus retrieved 113 papers. After filters (English, journal articles, 2019–2024, focus on wind forecasting), 73 remained. A detailed screening then selected those employing interval prediction. The topic modeling results were validated for consistency.

Research questions were defined to organize the review, addressing aspects from data origin to performance metrics, as summarized in Table 2. The methodological flow is illustrated in Figure 4.

The consolidation of literature reviews on wind power forecasting is essential because previous studies, although numerous, remain fragmented and mainly focused on specific techniques or point forecasting. The rapid evolution of hybrid models — combining modal decomposition methods (VMD, CEEMDAN, EMD), statistical learning, and deep neural architectures (LSTM, Gated Recurrent Unit (GRU), Convolutional Neural Network-Bidirectional Long Short-Term Memory (CNN-BiLSTM)) — has produced a growing yet dispersed body of research. While some reviews have addressed general aspects of wind forecasting or probabilistic prediction, few have specifically explored hybrid and interval forecasting approaches in an integrated manner. Therefore, this study seeks to synthesize and critically analyze these recent developments, providing a structured overview of methodological trends, current challenges, and future research directions. By consolidating this information, the review offers a comprehensive reference for researchers and practitioners aiming to enhance forecast reliability and support decision-making under uncertainty.

Point forecasting is defined as the prediction of a single expected value of wind power output for a given time horizon, typically representing the mean or median of the forecast distribution. Although this approach remains widely adopted due to its simplicity and interpretability, it does not capture the uncertainty associated with wind variability and is therefore often complemented by probabilistic or interval forecasting methods [21].

Hybrid models are defined as forecasting approaches that combine complementary techniques—such as modal decomposition methods (e.g., VMD, EMD, CEEMDAN), statistical learning, and deep neural networks (e.g., LSTM, GRU, CNN-BiLSTM)—to leverage their respective strengths in capturing temporal, nonlinear, and stochastic characteristics of wind power. By integrating signal decomposition with machine learning, hybrid models improve both the accuracy and robustness of forecasts across different time scales and operational conditions [22].

Finally, support for decision-making under uncertainty refers to the use of forecasting methods that explicitly account for the stochastic nature of wind generation. In this context, the integration of interval and hybrid forecasting frameworks provides essential tools for operators and policymakers to quantify risks, optimize power scheduling, and enhance renewable integration, thereby improving system reliability and economic efficiency [18].

To facilitate readability, a comprehensive list of all acronyms and abbreviations employed in this study is provided in Table 3, included as Appendix.

## III. RESULTS

From the initial set of 73 articles, 9 were excluded—4 due to lack of open access and 5 for being outside the scope, as they did not address wind energy forecasting. Thus, 64 articles were retained and classified into four forecasting approaches:

1) Deterministic Point-to-Point Forecasting: Estimates single values without considering uncertainty, e.g., wind speed at a given time [23];
2) Probabilistic and Interval Forecasting: Provides value ranges with confidence levels, capturing uncertainty [24];
3) Deterministic Interval Forecast: Defines fixed intervals by deterministic rules, without probabilistic meaning [25];
4) Hybrid Deterministic Forecasting: Combines deterministic methods to improve accuracy, e.g., statistical and machine learning techniques [23].

These categories guided the analysis of predictive approaches according to their objectives. Articles on Deterministic Interval and Probabilistic/Interval Forecasting were prioritized, as this work focuses on interval forecasting models for wind power. These techniques are essential for accounting for variability and uncertainty in generation, supporting decision-making and operational efficiency.

Of the 64 articles, 17 addressed Probabilistic and Interval Forecasting and 8 focused on Deterministic Interval Forecasting, totaling 25 directly related to interval methods. Many of these employed hybrid techniques—combining statistical models, machine learning, and signal decomposition—to improve accuracy. Notably, VMD, LSTM, and neural

**TABLE 1.** Table of topics with titles, keywords, search terms and number of papers.

| Topic | Title | Keywords | Search Term | Number of Papers |
|---|---|---|---|---|
| Topic 1 | Model with Neural Networks and Interval Decomposition for Wind Speed Prediction | speed, prediction, lstm, multi, decomposition, neural, learning, term, network, interval | "wind speed" AND "prediction" AND ("machine learning" OR "forecasting model" OR "statistical model" OR "neural network") AND "interval" AND "decomposition" | 36 |
| Topic 2 | Hybrid Model and Vector Decomposition for Predicting Wind Speed and Efficiency | speed, prediction, lstm, multi, decomposition, neural, learning, term, network, interval | "decomposition" AND "wind speed" AND "prediction" AND "algorithm" AND "hybrid" AND "energy" AND ("forecasting method" OR "prediction model") | 30 |
| Topic 3 | Hybrid System for Wind Power Generation Forecasting with ARIMA Models and Machine Learning Techniques | speed, energy, arima, electricity, learning, generation, hybrid, system, forecast, forecasting | "wind speed" AND "energy" AND "arima" AND "generation" AND "hybrid" AND ("forecast" OR "forecasting") | 9 |

**TABLE 2.** Overview of research questions and motivations.

| Question | Motivation |
|---|---|
| What is the origin of the data used in the studies reviewed? | To identify the main regions of the data sources used to forecast wind power generation, providing a solid basis for understanding trends and applications in the energy sector. |
| What is the origin of the publications analyzed? | Examine the countries of origin of the research and contributions to the field of renewable energy. |
| How is the publication of the studies reviewed distributed over time? | To evaluate the evolution and growth of research in the area over time, identifying periods of greater academic interest. |
| What are the prediction times evaluated in the studies? | To understand whether studies focus on short-, medium- or long-term predictions, helping to identify specific gaps and areas of opportunity. |
| What pre-processing techniques were used? | Explore the data preparation steps in the studies analyzed, analyzing the techniques used to pre-process the data. |
| Which prediction techniques were used? | Map the methods used for interval forecasts, highlighting hybrid techniques and interval models. |
| What evaluation metrics were used? | Identify the metrics used to evaluate the effectiveness of the models, such as RMSE, MAE, Interval Score, among others, and understand their suitability for measuring the performance of prediction techniques. |
| What are the proposals for future work in the area? | To provide a consolidated view of the gaps and directions suggested by the studies analyzed, such as the exploration of new hybrid techniques or improvements in the performance of interval models. |

network-based models were frequent, proving effective for nonlinear and variable wind data.

Results also revealed a growing use of probabilistic and interval methods, reinforcing their importance for wind power forecasting in high-uncertainty contexts and highlighting the trend toward robust and comprehensive approaches aligned with sustainability and energy efficiency goals.

### *A. ORIGIN OF DATA AND PUBLICATIONS*

Figure 5 shows the origin of the data used in the research analyzed. It can be seen that China is the main supplier of data for the studies, with approximately 15 articles using data from the country. Similarly, Figure 6 illustrates that China also leads as the place of publication of the articles, highlighting its relevance in the field of wind energy forecasting. In contrast, countries such as Germany, the USA and India are less representative, both as the origin of the data and as the place of publication.

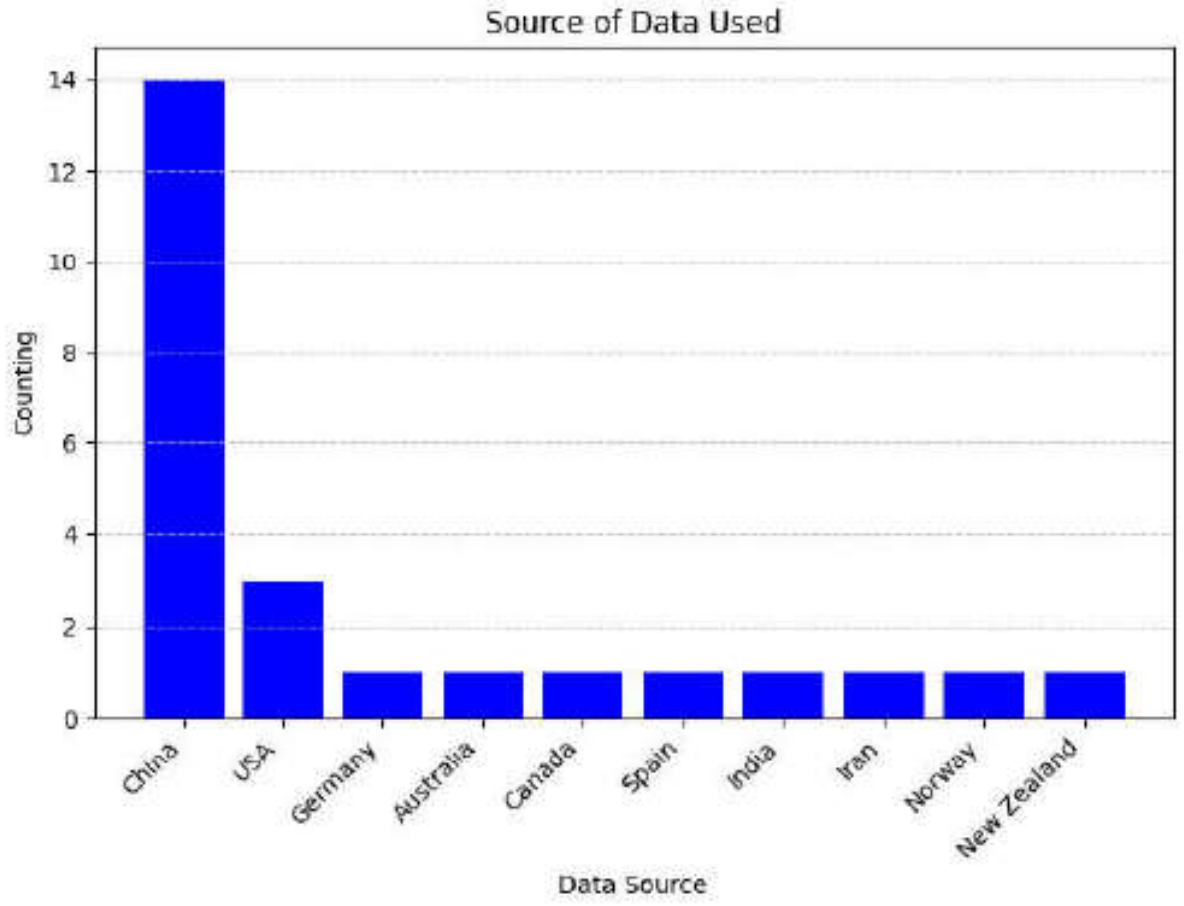

**FIGURE 5.** Origin of the data used in the research.

### *B. TEMPORAL DISTRIBUTION OF PUBLICATIONS*

The temporal distribution of publications is illustrated in Figure 7. Analyzing the data, there was a significant increase in the number of papers published in 2020 and 2022, with peaks of 7 publications per year. On the other hand, there was a reduction in 2023, followed by stabilization in 2024, with around 4 articles published annually. These papers focus mainly on interval predictive models for wind power generation.

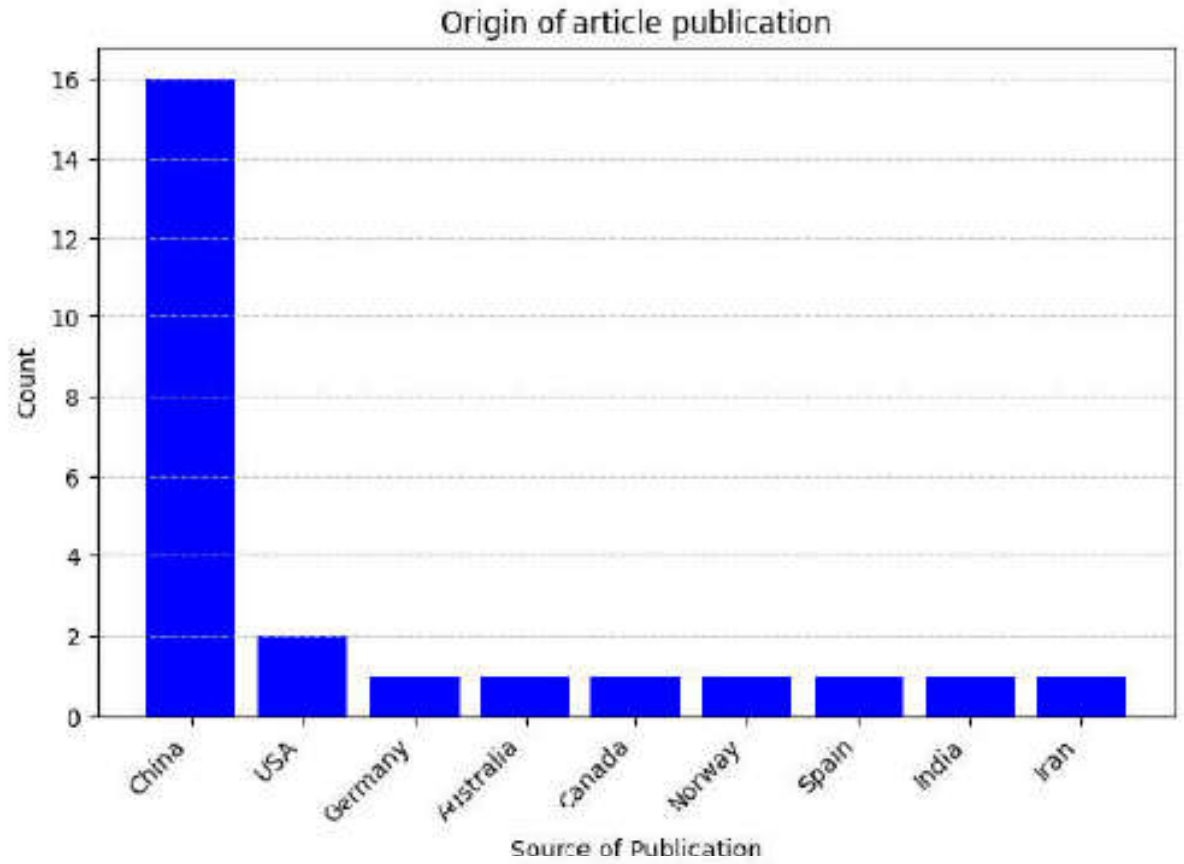

**FIGURE 6. Origin of the published articles.**

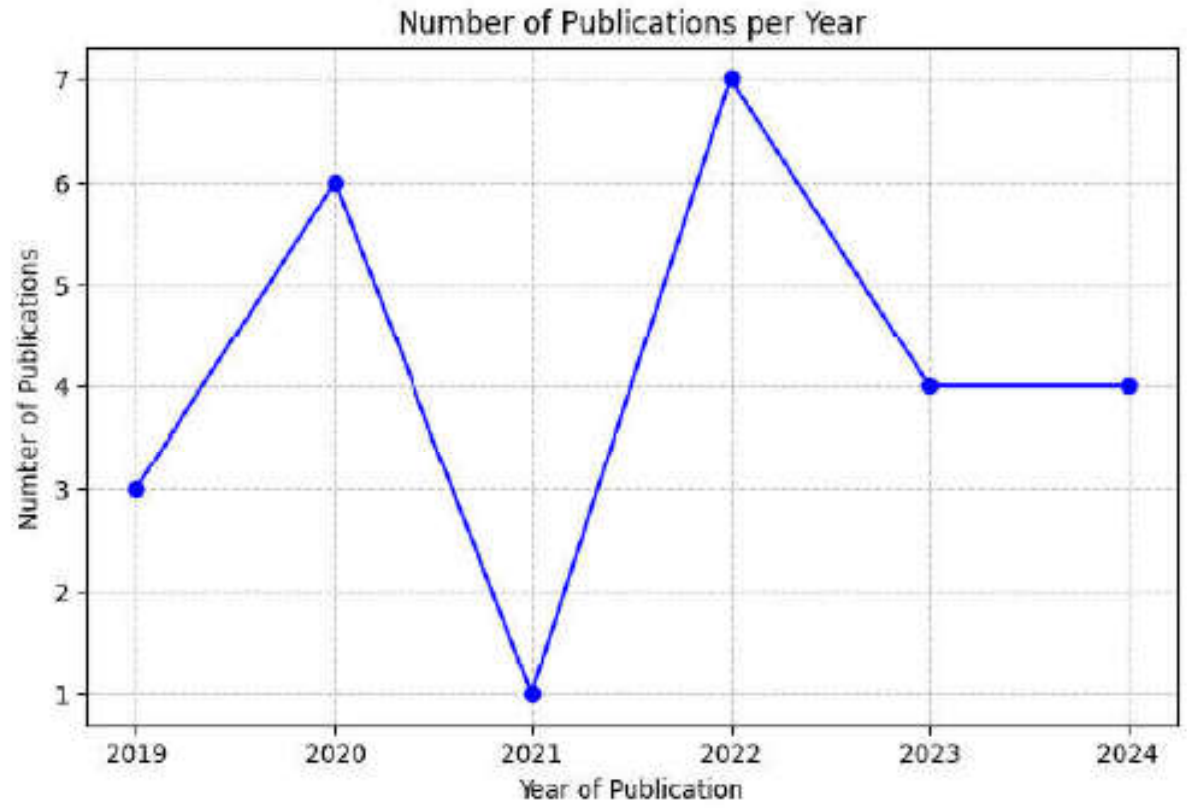

**FIGURE 7. Number of publications per year (2019-2024).**

## C. FORECASTING HORIZON

The analysis of forecasting horizon is shown in Figure 8. The vast majority of studies (96%) focus on short-term forecasts, highlighting the relevance of immediate solutions for the daily operation of wind farms. Only 4% of the studies dealt with medium-term forecasts, highlighting a gap in research into long-term energy planning.

## D. PRE-PROCESSING

Among the 25 publications analyzed, the most frequently used methods were VMD, CEEMDAN, and EMD, which stood out for their widespread adoption across the reviewed literature.

The VMD technique was the most frequent, applied in nine articles. CEEMDAN appeared in seven articles and EMD in five. Other relevant techniques, such as Singular Spectrum Analysis (SSA), Refined Composite Multiscale Fuzzy Entropy (RCMFE), Self-Organizing Map (SOM) and Wavelet Soft Threshold (WST), were used less frequently, but played specific roles, such as smoothing and noise removal. In addition, 10 of the studies analyzed adopted hybrid approaches, combining more than one technique simultaneously in data pre-processing.

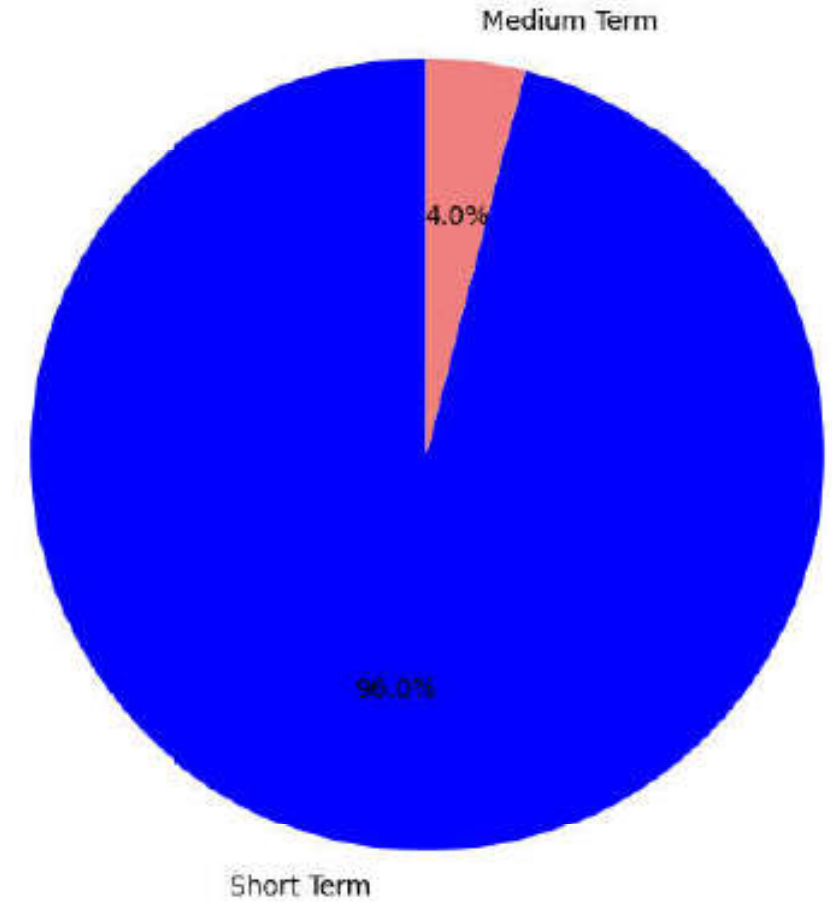

**FIGURE 8. Distribution of forecasting horizon in the analyzed publications.**

## E. FORECASTING METHODS

Traditional methods, particularly the Auto-Regressive Integrated Moving Average (ARIMA) model, remain widely used across several studies—such as those by Blei et al. [19], Gan and Qi [26], Bazionis and Georgilakis [23], and Gneiting and Katzfuss [24]—underscoring the enduring relevance and robustness of ARIMA in time series forecasting tasks.

It can be seen that traditional techniques, such as the ARIMA, continue to be widely used in several papers [27], [28], [29], [30], demonstrating their robustness for time series. More recent and sophisticated methods, such as LSTM [31], [32], [33] and Extreme Learning Machine (ELM) [34], [35], [36], [37], have also been highlighted in the literature due to their ability to deal with non-linear and high-dimensional data. Additionally, interval techniques, such as Kernel Density Estimation [35], [38], [39], [40], and hybrid methods, such as CNN-BiLSTM [41], demonstrate a growing trend in the field of wind power generation forecasting.

These results highlight the diversity of approaches explored by researchers and reflect an ongoing effort to integrate classical and contemporary methods in order to improve the accuracy and reliability of forecasts.

## F. PERFORMANCE MEASURES

Analysis of the performance measures used in the reviewed studies reveals a clear overview of the predominant approaches to evaluating wind energy forecasts.

RMSE, MAE, and MAPE stand out as the most widely adopted performance measures, reflecting the predominance of evaluations based on absolute accuracy and minimizing deviations between forecasts and actual values. The RMSE, by its nature of penalizing larger errors, is often used to capture the magnitude of extreme deviations, while the MAE provides a more balanced view of the model's average accuracy. MAPE, in turn, expresses the forecast errors as a percentage, which is particularly useful when comparing the performance of the model between different datasets or units [42]. This combined approach highlights a recurring concern among researchers to assess model performance both in average terms and in scenarios of greater error. These performance measures are defined as follows:

$$\text{RMSE} = \sqrt{\frac{1}{n}\sum_{i=1}^{n}(y_i - \hat{y}_i)^2}, \tag{1}$$

$$\text{MAE} = \frac{1}{n}\sum_{i=1}^{n}\left|y_i - \hat{y}_i\right|, \tag{2}$$

$$\text{MAPE} = \frac{100}{n}\sum_{i=1}^{n}\left|\frac{y_i - \hat{y}_i}{y_i}\right|. \tag{3}$$

In these formulas, $y_i$ represents the actual observed value, $\hat{y}_i$ is the value predicted by the model, and $n$ is the total number of observations. RMSE computes the square root of the average of squared errors, placing greater emphasis on larger errors. MAE calculates the average of absolute errors, providing a more uniform measure of the model's overall accuracy. MAPE computes the mean absolute percentage error, allowing for an intuitive understanding of forecast performance in relative terms.

When we look at performance measures specifically aimed at interval forecasting, such as the PICP and the CWC, we see a growing movement in the literature towards quantifying forecast uncertainty. The PICP measures the proportion of true values contained within the predictive intervals generated, while the CWC introduces a penalty for excessively wide intervals, ensuring a balance between coverage and accuracy [43]. These performance measures are fundamental for interval forecasts, as they allow us to assess the reliability and practical usefulness of the intervals generated, which is essential for decision-making in the energy sector. These performance measures are defined as follows:

$$\text{PICP} = \frac{1}{n}\sum_{i=1}^{n} c_i, \quad \text{where } c_i = \begin{cases} 1, & \text{if } y_i \in [\hat{y}_i^L, \hat{y}_i^U] \\ 0, & \text{otherwise,} \end{cases} \tag{4}$$

$$\text{CWC} = \text{NMPIW} \cdot (1 + \exp(-\eta(\text{PICP} - \mu))). \tag{5}$$

In these formulas, $y_i$ is the actual observed value, $\hat{y}_i^L$ and $\hat{y}_i^U$ are the lower and upper bounds of the prediction interval, respectively. The indicator $c_i$ is equal to 1 if the true value lies within the interval and 0 otherwise. PICP quantifies the empirical coverage of the prediction interval. The CWC (Coverage Width-based Criterion) includes the normalized mean prediction interval width (NMPIW), and adds an exponential penalty based on how much the actual coverage (PICP) falls below a predefined nominal confidence level $\mu$, weighted by a penalty coefficient $\eta$.

The use of performance measures such as the Continuous Ranked Probability Score (CRPS), which is still not widespread, highlights a gap in the incorporation of more sophisticated approaches to evaluating probabilistic forecasts. Unlike traditional performance measures, the CRPS takes into account the entire probability distribution of the forecast, providing a more detailed assessment of the model's performance under different uncertainty scenarios [44]. The CRPS is defined as follows:

$$\text{CRPS}(F, y) = \int_{-\infty}^{\infty} (F(z) - \mathbb{1}\{z \geq y\})^2 \, dz. \tag{6}$$

Here, $F(z)$ is the cumulative distribution function (CDF) of the forecast, $y$ is the observed value, and $\mathbb{1}\{z \geq y\}$ is the indicator function that equals 1 when $z \geq y$, and 0 otherwise. CRPS measures the squared distance between the forecast CDF and the step function defined by the actual observation, summarizing both calibration and sharpness of probabilistic forecasts.

Another relevant point is the presence of the Coefficient of Determination ($R^2$), which although not a central criterion in the evaluation of interval forecasts, is still used in some studies to measure the degree of explainability of models. However, its application in interval forecasts can be limited, since the $R^2$ was designed mainly to evaluate model adjustments in deterministic regressions. Its definition is given by:

$$R^2 = 1 - \frac{\sum_{i=1}^{n}(y_i - \hat{y}_i)^2}{\sum_{i=1}^{n}(y_i - \bar{y})^2}. \tag{7}$$

In this equation, $y_i$ is the actual value, $\hat{y}_i$ is the predicted value, and $\bar{y}$ is the mean of the actual values. The $R^2$ coefficient expresses the proportion of the variance in the dependent variable that is predictable from the independent variables.

## IV. DISCUSSION

The results confirm China's leading role in wind energy forecasting research, both as a primary data source and in publication output, likely driven by the country's strong investment in renewables and demand for advanced optimization technologies [34]. The temporal distribution of studies shows steady growth, with 2023 suggesting a possible stabilization or shift in focus. Recent works, such as [29], continue to prioritize short-term practical solutions while highlighting the potential of probabilistic models for broader horizons.

Short-term forecasting dominates the literature, reflecting the operational needs of wind farms for energy dispatch and grid stability. Such models benefit from lower uncertainty and adaptability to recent data, making them attractive for real-time applications. Wang et al. [32] and Li et al. [31] exemplify this trend by developing recurrent neural network architectures tailored to limited time windows.

Wang et al. [32] used 10-minute resolution data for a 16.5-hour horizon, addressing recurrent LSTM limitations such as one-step lag and proposing an online ultra-short-term forecasting platform. Li et al. [31] also applied 10-minute resolution data for two-day forecasts, validating multi-step predictions and introducing new cost functions for LSTM training to jointly optimize interval reliability and width.

Both studies reinforce that accurate wind speed forecasting is vital for efficient power system management, ensuring grid stability amid the growing penetration of intermittent renewable sources like wind power.

The predominance of short-term forecasts reflects the urgency of addressing operational challenges such as intermittency and variability, underscoring the need for fast and efficient solutions integrated into power grid operations. In contrast, the limited focus on medium-term forecasting reveals a gap in long-term planning strategies, as noted in [45], where most models remain tailored to short horizons.

Interval methods have proven effective for short-term applications in wind farm management, as they provide not only central predictions but also uncertainty ranges. Unlike point forecasts, interval forecasts convey both reliability and sharpness, commonly evaluated by PICP and PINAW. Well-calibrated intervals balance these dimensions, making them essential for decision-making under uncertainty [46]. However, their use in medium-term horizons is hindered by computational complexity, amplified uncertainty, and challenges in real-world validation [47].

Progress in this direction is illustrated by Zhang et al. [38], who combined Quantile Regression (QR) with KDE to forecast wind speed up to 200 hours ahead. Their model generated full probabilistic distributions, demonstrating stable performance beyond typical short-term windows and highlighting the feasibility of applying advanced statistical methods to extended horizons.

Strengthening medium-term interval forecasting could yield significant benefits for energy policy and system planning, including better coordination of supply and demand, reduced operational costs, and enhanced energy security. Greater emphasis on hybrid and probabilistic approaches is therefore a promising path to improve the sustainability and efficiency of the energy matrix. These findings reinforce the importance of expanding interval and probabilistic methods, alongside broader international collaboration, to address medium-term challenges in wind forecasting [31].

### A. PREPROCESSING

Among the preprocessing techniques reviewed, VMD emerged as the most prominent, appearing in 36% of the analyzed studies. This reflects its robustness and efficiency in decomposing complex time series into modal components across multiple frequency bands. VMD has proven particularly effective in capturing the non-linearity and variability inherent in wind speed data—two key characteristics frequently encountered in time series used for interval forecasting.

CEEMDAN also showed significant adoption, being used in 28% of the studies. It has established itself as a powerful method for noise reduction and for decomposing non-stationary signals into more stable subcomponents, providing a strong foundation for predictive modeling. An enhanced variant, Improved Complete Ensemble Empirical Mode Decomposition with Adaptive Noise (ICEEMDAN), appeared in 12% of the studies, further reinforcing the relevance of adaptive decomposition techniques in wind energy forecasting.

Although more traditional, EMD remains relevant, being applied in 20% of the reviewed studies. Its continued presence highlights its value as a straightforward and effective approach for extracting intrinsic modal components, especially in research contexts that favor methodological simplicity and reliability.

Other techniques, such as SSA (16%), Fuzzy Time Series (8%), and RCMFE, SOM, and WST (each in 4% of the studies), were less frequent but play specific roles in signal smoothing and noise attenuation. These methods are often integrated into hybrid configurations, complementing primary decomposition strategies.

The adoption of hybrid approaches in 40% of the reviewed studies underscores a growing recognition that combining techniques is essential to effectively address the complexities of wind speed time series—particularly their non-linearity, non-stationarity, and high variability.

Ultimately, the preference for advanced decomposition techniques such as VMD and CEEMDAN highlights the scientific community's ongoing effort to improve the accuracy and robustness of forecasting models. At the same time, the use of hybrid strategies reflects a broader commitment to optimizing model performance by leveraging the complementary strengths of multiple preprocessing methods.

Figure 9 illustrates the frequency with which each pre-processing technique was used in the reviewed studies. VMD, CEEMDAN, and EMD were the most commonly adopted techniques, followed by SSA and ICEEMDAN. The frequent use of these methods reflects their effectiveness in handling non-stationary and noisy wind speed data, especially in hybrid modeling scenarios.

These findings show that advanced signal decomposition and smoothing techniques are crucial for improving accuracy and robustness in forecasting models. Methods such as VMD, CEEMDAN, and EMD effectively handle non-stationary and highly variable wind speed series by decomposing them into stable, interpretable subcomponents. This enables models to work with cleaner inputs, enhancing learning and reducing prediction error.

The integration of these techniques into hybrid frameworks—combining statistical methods, deep learning, and entropy-based approaches—marks a shift toward adaptive modeling pipelines. Such approaches leverage multiple preprocessing strategies to address non-linearity, seasonality, and data irregularity. For example, VMD combined with LSTM networks aligns temporal decomposition with sequential learning, yielding strong results.

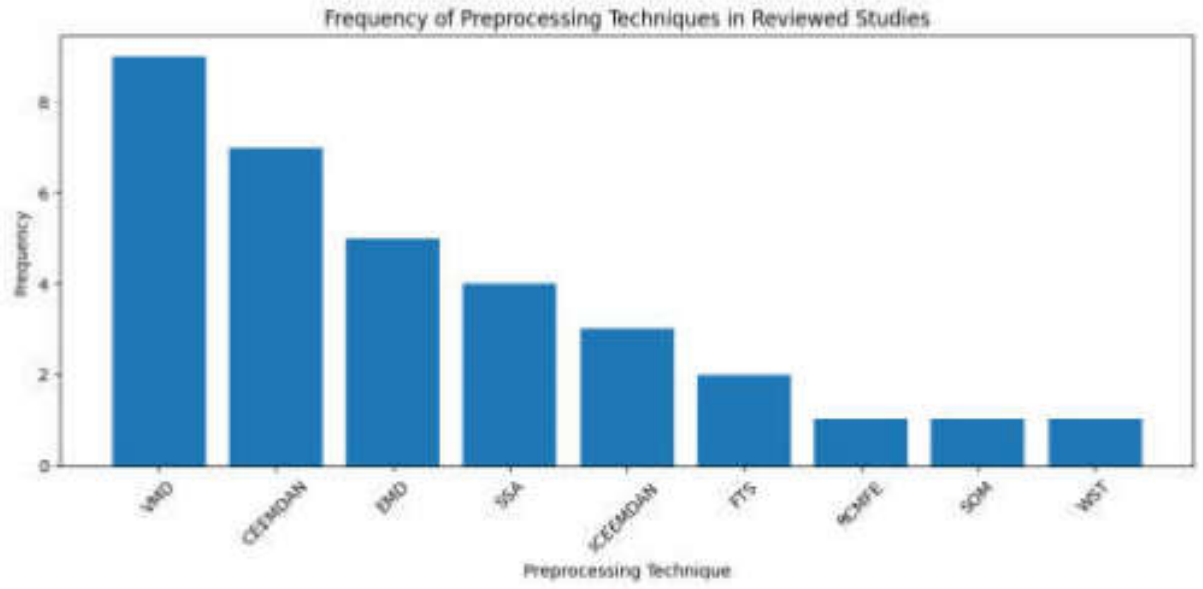

**FIGURE 9. Frequency of use of prediction techniques in the reviewed studies.**

Their widespread use in interval and probabilistic forecasting highlights their role in producing sharper, more reliable prediction intervals while maintaining coverage—an essential feature for wind energy applications, where decisions depend on both point estimates and forecast confidence.

Thus, the strategic use of multiple preprocessing steps, especially signal decomposition and smoothing, is not just a technical refinement but a foundation for building models that address the complexity and uncertainty of wind energy series.

### *B. FORECASTING METHODS*

The results of the review indicate a significant diversity of forecasting techniques applied to wind power generation, highlighting the evolution of the approaches used to model wind variability. The techniques analyzed range from traditional statistical models to hybrid approaches that combine deep learning, Quantile Regression and modal decomposition. This variety reflects the ongoing effort in the literature to improve the accuracy and reliability of forecasts, minimizing uncertainties and improving decision-making in the energy sector.

#### 1) TRADITIONAL MODELS AND THEIR LIMITATIONS

Classical statistical models, such as ARIMA, remain widely used for short-term forecasts due to their robustness in modeling time series and capturing cyclical patterns, as shown by Zhang et al. [27], Ding and Meng [28], Li et al. [29], and Kirthika et al. [30]. However, ARIMA assumes linearity and struggles with volatility and non-stationarity, leading researchers to combine it with modal decomposition [32], [48] or machine learning [33] to enhance predictive capacity.

Another statistical model present in the studies reviewed is QR, used to predict confidence intervals of estimates, providing more informative forecasts, as shown by Wang et al. [49] and Zhang et al. [38]. QR has proven to be a valuable technique when combined with deep learning-based methods, allowing the dispersion of data to be modeled in a more realistic way.

With technological advances and more data, Artificial Neural Networks, particularly LSTM and GRU, have become prominent alternatives [31], [32], [33], [45], excelling in wind time series prediction by modeling long-term dependencies. Advanced models such as CNN-BiLSTM and Bi-LSTM also capture spatial and temporal patterns, as in Zhao et al. [41] and Bommidi et al. [50]. CNN integration enhances feature extraction from high-dimensional data, though these models demand large labeled datasets and significant computational resources.

Another important aspect is the growing adoption of hybrid models that combine neural networks with statistical techniques. Studies show that integrating ELM with modal decomposition can improve the stability and accuracy of forecasts, as presented by Li et al. [34], Xu et al. [35], Abdoos et al. [36]. Models such as the Gaussian Process and XGBoost have also been explored as alternatives for refining forecasts based on deep learning, especially in the construction of interval forecasts, as shown by Wang et al. [51].

Beyond these recurrent and convolutional architectures, recent advancements in deep learning have introduced Transformer-based models as a powerful alternative for modeling complex temporal and nonlinear dynamics. Unlike recurrent networks, Transformers employ self-attention mechanisms that enable parallel processing and efficient learning of both short- and long-range dependencies. This design enhances scalability, accuracy, and interpretability—key attributes for renewable energy forecasting, where temporal variability and nonlinear interactions dominate [52], [53].

Geneva and Zabaras [52] demonstrated that Transformer models can effectively capture the dynamics of chaotic and multiscale physical systems. By integrating physics-informed embeddings and exploring the mathematical equivalence between self-attention and time-integration schemes, their approach achieved superior long-term forecasting performance compared to conventional recurrent architectures. Their findings highlighted that Transformers maintain physical consistency in dynamic modeling while accurately representing multiscale temporal dependencies—capabilities highly relevant to wind generation systems.

Building upon this foundation, the FLUID-GPT framework introduced by Yang et al. [53]. represents one of the first hybrid Transformer–CNN architectures for engineering flow systems. By combining GPT-2 with convolutional layers, the model predicted particle trajectories and surface-erosion profiles with remarkable accuracy, achieving a 54% reduction in mean-squared error and a 70% reduction in training time compared to BiLSTM-based approaches. Its success highlights the potential of GPT, style Transformers to process high-dimensional sequential data and capture spatial temporal interactions in dynamic physical systems—an ability that could be extended to renewable-energy forecasting applications.

Building upon these advances, Hussan et al. [54] conducted a comprehensive review of Transformer-based frameworks in renewable energy forecasting. Their analysis confirmed the superior capability of Transformers to capture long-term dependencies and complex spatio-temporal interactions compared with LSTM and GRU models. Moreover, they identified emerging hybrid and probabilistic Transformer architectures that integrate quantile-based learning, Gaussian mixture models, and copula-based post-processing to enhance uncertainty representation in interval and scenario forecasting.

Collectively, these recent studies establish a solid theoretical and practical foundation for the adoption of attention-based architectures as a next-generation paradigm for deterministic, interval, and probabilistic forecasting in renewable energy systems. However, it is noteworthy that none of the studies analyzed in the present review have yet applied Transformer-based approaches to wind power interval forecasting. This gap highlights a promising research direction for future work—where attention mechanisms could be integrated with modal decomposition and hybrid frameworks to further improve forecasting accuracy, computational efficiency, and uncertainty quantification.

Recent advancements in time series forecasting for energy systems have integrated deep learning, hybrid models, and signal decomposition to improve predictive accuracy and robustness. In hydroelectric power plants, Stefenon et al. [55] presented the Neural Hierarchical Interpolation Time Series model, tailored for multi-horizon reservoir level forecasting. The model captures hierarchical temporal patterns, enabling more accurate long-term forecasts for hydropower planning.

Building on deep learning, Stefenon et al. [56] proposed a hypertuned wavelet CNN with LSTM layers, combining wavelets' time-frequency localization with LSTMs' sequence learning. This achieved significant improvements in hydroelectric forecasting. Similarly, Stefenon et al. [57] developed a hypertuned Temporal Fusion Transformer, showing how attention mechanisms enhance dam level prediction by weighing temporal and exogenous inputs.

Signal decomposition also improves preprocessing. Larcher et al. [58] proposed a hybrid model using EMD and VMD to disentangle nonlinear patterns before applying Echo State Networks, improving streamflow forecasting.

In a broader energy context, da Silva et al. [59] compared machine learning models for hydroelectric inflow prediction in the Brazilian grid, highlighting hyperparameter optimization and ensemble methods as key for decision support.

Beyond hydropower, these methods apply to wind energy. Ribeiro et al. [60] combined VMD and Bagging Extreme Learning Machines, optimized via multi-objective strategies, improving wind power prediction. Moreno et al. [61] integrated SSA with VMD and machine learning, significantly reducing wind speed forecast errors.

Energy price forecasting has also advanced. Klaar et al. [62] applied ensemble learning and seasonal decomposition for Latin American markets, especially Mexico, showing the importance of structural optimization to capture market dynamics.

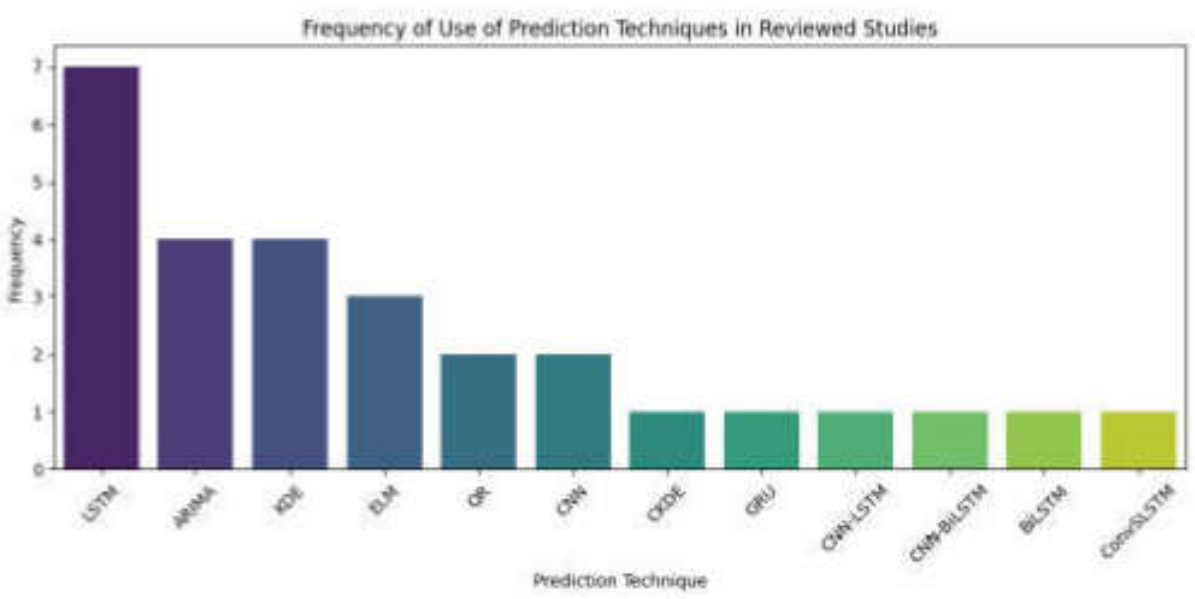

**FIGURE 10. Frequency of use of prediction techniques in the reviewed studies.**

Machine learning is also applied in predictive maintenance. Stefenon et al. [63] investigated insulator fault prediction with the Group Method of Data Handling and the Christiano-Fitzgerald filter, illustrating how tailored filtering and optimization improve grid reliability.

### 2) INTERVAL AND PROBABILISTIC TECHNIQUES

Interval forecasting has become increasingly relevant for addressing the variability and uncertainty of wind generation. Notable approaches include KDE and Conditional Kernel Density Estimation (CKDE), which estimate probabilistic forecast distributions and enable more robust confidence intervals, as shown by Jiang et al. [64], Zhang et al. [38], He et al. [39], Xu et al. [35], and Wang et al. [40].

Monte Carlo Markov Chain (MCMC) and Twin Support Vector Regression were also applied in interval forecasting, as reported by Li et al. [29], and Fu et al. [46], highlighting efforts to integrate probabilistic approaches that enhance uncertainty modeling and support practical applications in energy planning and wind farm operation.

Additionally, CRPS has been used as a criterion to evaluate probabilistic forecasts, as demonstrated by He et al. [39], Abdoos et al. [36], and Zhao et al. [37], demonstrating a transition in the literature towards approaches that consider not only point accuracy, but also the quality of forecast intervals.

Figure 10 shows the frequency of prediction techniques across the reviewed studies. LSTM is the most widely used method, particularly within hybrid models, reflecting the strong trend toward machine learning approaches. In the figure, the LSTM and CNN columns include all hybrid models that employ these techniques, while hybrids are also shown separately in the last columns. Besides hybrids, the most common standalone methods are ARIMA, KDE, ELM, and LSTM.

In addition, Figure 11 presents a heatmap linking prediction techniques with performance measures. It highlights the dominance of point-based accuracy criteria (RMSE, MAE) across most methods, while also showing that models based on QR, KDE, and deep hybrid architectures adopt interval metrics (PICP, CWC, CRPS), reflecting

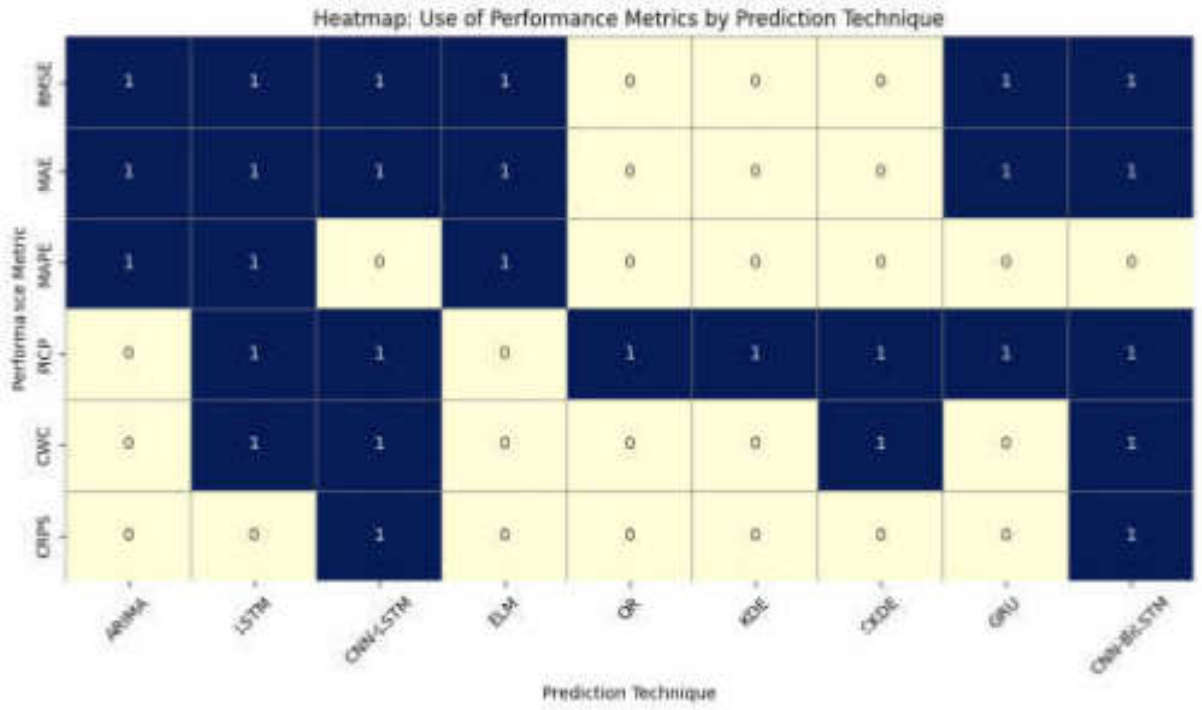

**FIGURE 11. Heatmap showing the use of performance measures by forecasting approach.**

increasing attention to forecast uncertainty and probabilistic modeling.

These figures illustrate how predictive approaches align with evaluation criteria in wind energy forecasting, emphasizing the methodological diversity in the literature. The review also highlights the growing use of hybrid models, which combine modal decomposition, deep learning, and QR to capture both data dynamics and forecast uncertainty.

VMD and CEEMDAN have become standard techniques for decomposing wind time series into frequency components, as shown by Jiang et al. [64], Li et al. [31], and Li et al. [29]. These methods have been enhanced and applied in diverse contexts by Sun and Jin [48], Wang et al. [32], and Zhao et al. [41]. Their integration with statistical and neural models has further improved forecasting quality, as seen in He et al. [39], Kirthika et al. [30], Xu et al. [35], and Li et al. [65].

In addition, CNN-LSTM and CNN-BiLSTM models have also been explored, notably by Zhao et al. [41], and Li et al. [65], combining the ability of LSTMs to capture temporal patterns with the efficiency of CNNs in feature extraction. However, their high computational cost remains a challenge, limiting applicability in operational settings with constrained resources. Hybrid models have gained traction as they combine complementary techniques—such as statistical models, machine learning algorithms, and signal decomposition methods—to improve the reliability and precision of wind power forecasts. One prominent strategy is the integration of VMD with recurrent neural networks like LSTM, which enables the capture of nonlinear and dynamic temporal structures in wind data. This approach has shown considerable promise in enhancing short-term interval forecasts, as demonstrated by Li et al. [31] and Wang et al. [32], where VMD-LSTM frameworks produced more accurate and stable prediction intervals.

The effectiveness of hybrid models is further amplified when combined with advanced signal decomposition methods such as CEEMDAN. Studies by Zhao et al. [41] and Xu et al. [35] highlight that CEEMDAN-based hybrid frameworks not only reduce interval width but also increase coverage probability, balancing the trade-off between sharpness and reliability in interval forecasting.

In addition, interval-based models continue to demonstrate their importance in representing the uncertainty inherent in wind power generation. By estimating a range of possible outcomes—rather than a single value—these models offer a more informative basis for decision-making. Techniques such as KDE, QR, and MCMC have been widely adopted to construct probabilistic intervals. For instance, Zhang et al. [38] proposed a KDE-QR-based model capable of delivering forecasts up to 200 hours ahead, indicating the growing applicability of interval forecasting in extended time horizons.

These results collectively emphasize the relevance of hybrid interval models in operational contexts, as they simultaneously enhance forecast accuracy and provide robust uncertainty quantification—two essential aspects for reliable energy planning and grid management in the presence of high wind variability.

### 3) CHALLENGES AND OPPORTUNITIES

Although the techniques reviewed have shown considerable progress in wind energy forecasting, certain challenges persist. One key issue is the lack of standardization in the selection of performance measures, which complicates direct comparisons between models. While applying different forecasting approaches to a common dataset is a technically feasible task—often requiring only the implementation and execution of existing algorithms—the literature still lacks systematic benchmarking efforts that would enable more robust and reproducible comparisons. Furthermore, the limited number of studies assessing real-time applicability leaves open questions regarding the practical integration of these models into operational energy management systems, particularly in scenarios with dynamic data flows and infrastructure constraints

The trend observed in the studies reviewed suggests that interval forecasting will continue to evolve, incorporating new hybrid techniques and improving uncertainty modeling. The integration of probabilistic models with deep learning and modal decomposition appears to be a promising way to improve the reliability of forecasts and optimize the operation of wind power generation systems.

## C. PERFORMANCE MEASURES

The results indicate that wind energy forecasting is still largely assessed by point accuracy, with RMSE and MAE being the most frequent metrics. Point accuracy refers to the proximity between predicted and observed values, assuming that the best model is the one with the smallest average error [66]. Figure 12 shows the frequency of performance measures in the analyzed studies, distinguishing between point and interval forecasts and by model complexity.

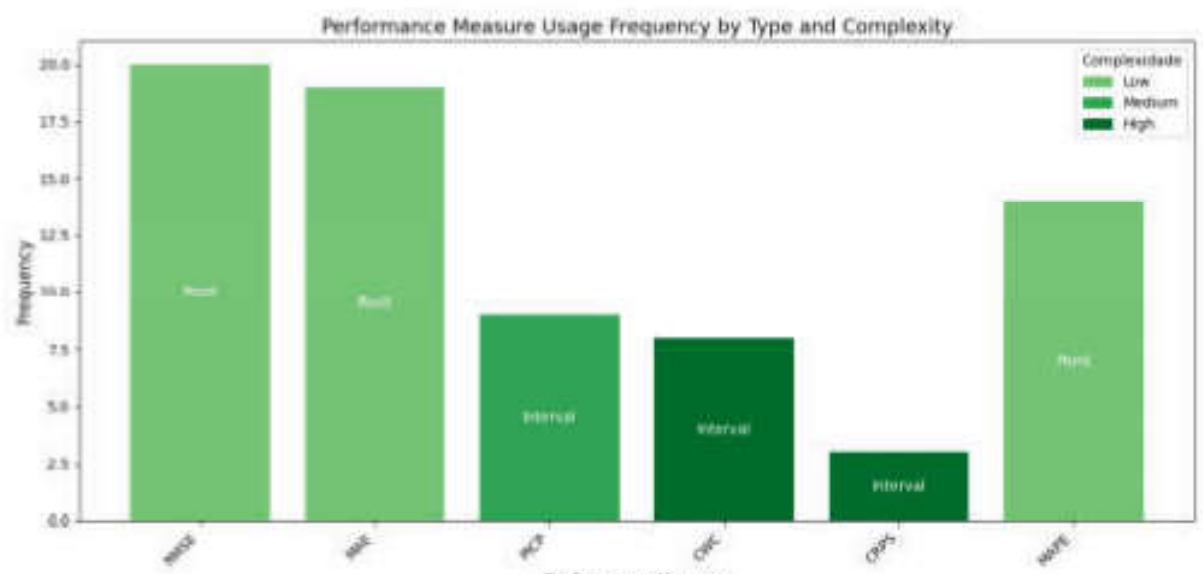

**FIGURE 12.** **Frequency of usage of performance measures, categorized by forecast type and model complexity.**

Although this focus reflects a tradition in time series forecasting, it is insufficient for real scenarios where wind variability requires reliability assessment. The growing use of interval measures such as PICP and CWC reflects a shift toward more informative forecasts. In this context, Prediction Interval Normalized Average Width (PINAW) and Prediction Interval Normalized Root-mean-square Width (PINRW) are also relevant, since they evaluate the sharpness of intervals by penalizing excessive width while keeping comparability across datasets. More robust distribution-based metrics such as CRPS remain rare, partly due to computational complexity and the demands of deep learning and modal decomposition models.

The variety of measures used also complicates comparisons across studies. Traditional approaches such as ARIMA are usually evaluated with RMSE and MAE, whereas probabilistic and interval methods (e.g., KDE, QR) more often employ PICP, CWC, and width-based criteria such as PINAW or PINRW. Signal decomposition (VMD, CEEMDAN) appears in both contexts, improving data quality; when combined with deep learning (CNN-BiLSTM, LSTM), these methods support a wider choice of evaluation metrics. Nonetheless, even advanced models such as Bi-LSTM and CNN-BiLSTM are sometimes assessed only with RMSE and MAE, showing a misalignment between model complexity and evaluation.

In practice, forecasts based solely on RMSE and MAE may be accurate but fail to quantify uncertainty. Interval-based measures, by contrast, provide confidence ranges that support more efficient wind power integration into the grid. Future work should strengthen the use of advanced probabilistic measures (e.g., CRPS) and ensure coherence between evaluation metrics and modeling techniques, avoiding misleading comparisons and improving the reliability of wind forecasting.

In addition to the analysis of performance measures, it is essential to understand how the models construct the bounds of the prediction intervals. Accordingly, the next section examines in detail the 25 selected articles, highlighting the strategies adopted to define the lower and upper bounds and assessing their ability to represent the uncertainty associated with wind power forecasting.

## V. GENERAL INTRODUCTION TO INTERVAL FORECASTING FORMULATIONS

### *A. RECURRENT FORMULATIONS OF PREDICTION INTERVALS*

Across the set of reviewed articles, two main formulations for constructing prediction intervals are recurrent:

1) Symmetric Interval Based on the Normal Distribution of Residuals

This approach is employed in **Articles 1, 5, 8, and 21**, where residuals are assumed to follow a normal distribution. The prediction interval is then derived from the point forecast and the standard deviation of the residuals, using the quantile of the standard normal distribution corresponding to a given confidence level:

$$PI(t) = \left[\hat{y}_t - z_{\alpha/2} \cdot \sigma_t,\ \hat{y}_t + z_{\alpha/2} \cdot \sigma_t\right], \quad (8)$$

where:

- $\hat{y}_t$ is the point forecast at time $t$;
- $\sigma_t$ is the estimated standard deviation of the residuals;
- $z_{\alpha/2}$ is the quantile of the standard normal distribution corresponding to the desired confidence level.

2) Asymmetric Interval Based on Quantiles

This approach is used in **Articles 3, 4, and 9**, where the prediction interval is constructed directly from the estimation of quantiles of the predictive distribution, without assuming symmetry or normality. Quantiles can be estimated using methods such as KDE, QR, or ensemble models:

$$PI(t) = \left[Q_{\alpha/2}(t),\ Q_{1-\alpha/2}(t)\right], \quad (9)$$

where $Q_{\alpha/2}(t)$ and $Q_{1-\alpha/2}(t)$ are, respectively, the lower and upper quantiles of the conditional distribution of the predicted variable at time $t$.

In addition to these two recurrent formulations, some studies employ specific approaches that do not fit directly into either category. Examples include Monte Carlo methods [28], [36], Gaussian Process Regression (GPR) [51], MCMC [29]. These methods will be addressed individually in subsequent sections, given the particular nature of their probabilistic estimates.

#### 1) A1 - JIANG ET AL. [64]

In this study, the probabilistic prediction interval was generated from the predictive distributions obtained for each reconstructed sub-series of the original series, using the CKDE-NRC method (Conditional Kernel Density Estimation with bandwidth determined by the normal reference criterion). After decomposing the data through Refined Variational Mode Decomposition, the probability density of each resulting sub-series was individually predicted. The variance of the overall forecast at time $n + 1$ was then calculated by considering the sum of the predicted variances of the sub-series, along with the covariances among them. Based on this total variance and the predicted mean, and assuming that the target variable (wind speed) follows a normal distribution, the prediction interval was constructed using the quantile of the standard normal distribution corresponding to the established confidence level. Thus, the interval is defined as follows:

$$PI(n+1) = 2\sigma(n+1) \cdot U_{\alpha/2}, \tag{10}$$

where:

- $PI(n+1)$: width of the prediction interval at time $n+1$;
- $\sigma(n+1)$: total standard deviation of the forecast at time $n+1$, obtained from the aggregated variance of the reconstructed sub-series;
- $U_{\alpha/2}$: quantile of the standard normal distribution corresponding to the confidence level $1-\alpha$;
- $\alpha$: significance level (e.g., $\alpha = 0.05$ for a 95% confidence interval).

The experiments showed that the method produced intervals with high coverage (PICP $\geq$ 96%) and reduced width (PINAW $\leq$ 69%), maintaining a balance between accuracy and reliability, with superior performance compared to benchmark models.

#### 2) A2 - LI ET AL. [34]

In this work, a hybrid system named ICEEMDAN-F-MOICA-ELM is proposed, which combines time series decomposition, multi-objective optimization, and machine learning to generate both point and interval forecasts. The intervals are constructed in a non-parametric manner, aiming simultaneously to maximize coverage Interval Forecast Coverage Probability (IFCP) and minimize width Interval Forecast Normalized Average Width(IFNAW). The prediction interval is defined based on the optimization of two criteria:

- Maximize IFCP, ensuring that the largest possible proportion of actual values is contained within the predicted intervals;
- Minimize IFNAW, in order to obtain narrower and more informative intervals.

The MOICA algorithm balances coverage and accuracy, producing adaptive intervals. The ICEEMDAN-F-MOICA-ELM model outperformed the ICEEMDAN-ICA-ELM benchmark, achieving, for a width coefficient of 0.1, gains of 14.60% in IFCP and 1.54% in IFNAW; and for a coefficient of 0.2, improvements of 11.17% in IFCP and 3.67% in IFNAW. Thus, the proposed method provides more reliable and informative intervals for wind power forecasting.

#### 3) A3 - JIANG ET AL. [47]

In this study, the interval forecast was obtained through CKDE, which estimates the conditional density of wind speed from the time series history, without assuming a parametric distribution of the residuals. The methodology consisted of: (i) decomposing the series using the Robust Local Mean Decomposition (RLMD) method; (ii) generating point forecasts for each component with the Group Method of Data Handling (GMDH); and (iii) reconstructing the series and applying CKDE to extract the corresponding quantiles (90%, 95%, 99%), thus forming the prediction interval limits. The prediction interval at time $n+1$ is defined as:

$$PI(n+1) = [F^{-1}(\alpha/2), F^{-1}(1-\alpha/2)], \tag{11}$$

where:

- $F^{-1}$: quantile function (inverse of the Cumulative Distribution Function (CDF)) of the conditional density estimated by CKDE;
- $\alpha$: significance level (for example, $\alpha = 0.05$ for a 95% confidence interval).

The RLMD-GMDH-CKDE method achieved high performance, with a PICP of 96.5% (PINAW 44.2%) at 90% confidence, 98.5% (52.5%) at 95%, and 99.5% (69.1%) at 99%, outperforming ARIMA, Backpropagation Neural Network, and Support Vector Machine (SVM) by providing more informative and consistent intervals.

#### 4) A4 - XU AND YANG [67]

In this study, the interval forecast is constructed based on the forecast errors, defined by the following expression:

$$e = V_{\text{meas}} - V_{\text{pred}}, \tag{12}$$

where $e$ represents the forecast error, $V_{\text{meas}}$ is the actual measured wind speed, and $V_{\text{pred}}$ is the point forecast provided by the EMD-MKL model. To address the variability of errors (heteroscedasticity), the authors divide the range of possible wind speed values into subintervals $D_i$, grouping the errors according to the forecast range. In each subinterval, a kernel density estimation (KDE) is applied to model the error distribution in a non-parametric way:

$$f(e) = \frac{1}{Nh} \sum_{j=1}^{N} K\left(\frac{e - e_j}{h}\right), \tag{13}$$

where $f(e)$ is the estimated probability density function of the errors, $e_j$ are the sampled historical errors, $N$ is the total number of samples in the considered subinterval, $h$ is the bandwidth parameter that controls the smoothness of the estimate, and $K(\cdot)$ is the kernel function, assumed to be Gaussian in this case. From $f(e)$, the authors numerically integrate to obtain the cumulative distribution function $F(\xi)$, and then determine the lower and upper quantiles associated with the desired confidence level. Finally, the prediction interval for time $n+1$ is constructed by adding these quantiles to the point forecast:

$$PI(n+1) = \left[V_{\text{pred}} + F^{-1}(\alpha/2),\ V_{\text{pred}} + F^{-1}(1-\alpha/2)\right], \tag{14}$$

where $PI(n+1)$ represents the prediction interval at time $n+1$, $F^{-1}(\cdot)$ is the inverse function of the cumulative distribution of errors (i.e., the quantiles), and $\alpha$ is the significance level (for example, $\alpha = 0.10$ for a 90% confidence interval). In this way, the interval limits are defined directly from the empirical distribution of historical errors, adjusted to the predicted wind speed range, providing an interval forecast that accounts for model uncertainty.

### 5) A5 - ZHANG ET AL. [27]

In this research, an interval forecasting method for wind speed is proposed based on the Lorenz Disturbance Distribution (LDD), which incorporates the randomness and chaotic behavior of forecast errors without assuming classical distributions such as the normal. The model uses an ELM neural network for point forecasting, and the obtained residuals are employed to generate a probability distribution from the Lorenz dynamic system, which defines the prediction intervals.

The forecast error is defined as:

$$e_t = y_t - \hat{y}_t, \tag{15}$$

where $e_t$ represents the error at time $t$, $y_t$ is the actual wind speed value, and $\hat{y}_t$ is the value predicted by the ELM.

To model the error distribution, the authors propose the LDD density function, defined as:

$$L(x) = \frac{1}{Z} \exp\left(-\frac{(x-\mu)^2}{2\sigma^2 + \delta|x-\mu|}\right), \tag{16}$$

where:

- $x$: forecast error values;
- $\mu$: mean of the errors;
- $\sigma^2$: variance of the errors;
- $\delta$: disturbance parameter, adjusted based on the Lorenz system dynamics;
- $Z$: density normalization constant.

From this distribution, the authors extract the quantiles corresponding to the desired confidence level. The prediction interval with confidence level $1-\alpha$ is defined as:

$$PI(t) = \left[\hat{y}_t + q_{\alpha/2},\ \hat{y}_t + q_{1-\alpha/2}\right], \tag{17}$$

where:

- $PI(t)$: prediction interval at time $t$;
- $\hat{y}_t$: point forecast from the ELM;
- $q_{\alpha/2},\ q_{1-\alpha/2}$: lower and upper quantiles of the LDD distribution.

In addition to the main proposal, comparative tests were conducted with Bootstrap, QR, and QR+Bootstrap. The LDD model achieved superior performance in terms of PICP and PINAW at confidence levels of 80%, 90%, and 95%, confirming its effectiveness in representing uncertainty.

### 6) A6 - LI ET AL. [45]

This work incorporates interval forecasting into a hybrid model with GRU, VMD, and error correction, aiming to generate reliable and accurate intervals. The performance is evaluated using the metrics PICP, PINRW, and the CWC.

The accuracy of the interval is evaluated by the PINRW metric, which measures the average width of the intervals, normalized by the total range of the series:

$$\text{PINRW} = \frac{1}{nR}\sum_{i=1}^{n}(U_i - L_i)^2, \tag{18}$$

where $R = \max(x) - \min(x)$ represents the range of the wind speed series $x$.

To combine the two metrics, PICP and PINRW, into a single quality criterion for the interval, the authors apply a variant of the CWC, in which the NMPIW term of the classical formulation is replaced by PINRW:

$$\text{CWC}_{PINRW} = \text{PINRW}\left(1 + \gamma e^{-\eta(\text{PICP}-\mu)}\right), \tag{19}$$

$$\gamma = \begin{cases} 0, & \text{if PICP} \geq \mu \\ 1, & \text{if PICP} < \mu \end{cases}, \tag{20}$$

where $\mu$ represents the nominal confidence level (for example, $\mu = 0.90$), and $\eta$ is a hyperparameter that controls the penalty for PICP values lower than $\mu$.

The initial construction of the prediction bounds is based on the input series $x$, with an adjustable width coefficient $R_w$:

$$UB_{con} = x + R_w, \quad LB_{con} = x - R_w, \tag{21}$$

$$R_w = w \cdot (\max(x) - \min(x)), \quad w \in [0, 1]. \tag{22}$$

Next, the series is decomposed into modal components using VMD. The primary mode (the one with the highest energy) is used to construct the main bounds, while the residual modes are used to predict deviations. The GRU network is trained to learn the predicted bounds $UB_{pre}$ and $LB_{pre}$. To correct the deviations between the predicted bounds and the constructive bounds, the authors define the forecast errors:

$$UB_{Err} = UB_{pre} - UB_{con}, \quad LB_{Err} = LB_{pre} - LB_{con}. \tag{23}$$

These errors are then modeled by a second GRU (called EPM), which forecasts the values $UB_{Errpre}$ and $LB_{Errpre}$, used to calibrate the final bounds:

$$UB_{cal} = UB_{pre} - UB_{Errpre}, \quad LB_{cal} = LB_{pre} - LB_{Errpre}. \tag{24}$$

The VMD-GRU-EC model achieved PICP > 96% and outperformed SVM, KELM, and ANN in terms of PICP, PINRW, and CWC, demonstrating greater effectiveness in generating reliable intervals for short-term wind speed forecasting.

### 7) A7 - DING AND MENG [28]

In this work, interval (or probabilistic) forecasting is implemented by the Probability Interval Prediction Module, as described in Section III-C of the article. The objective of this module is to construct PIs for the wind speed series, providing an explicit measure of the uncertainty associated with the forecasts. The methodology combines ARIMA, for deterministic forecasting, and Improved First-Order Markov Chain (IFOMC), which is responsible for constructing the prediction intervals.

In the preprocessing stage, the series is divided into two parts: cline (linear), predicted by ARIMA, and cremaining (nonlinear), used by IFOMC to generate the intervals. The interval forecasting process is carried out in four steps:

- (1) **State Categorization:** the cremaining series is discretized into $k$ states in an equal-frequency manner, so that each state contains approximately the same number of observations, avoiding the sparsity typical of traditional Markov chains.
- (2) **Construction of the Transition Matrix:** based on the obtained state series, the transition matrix $P = [p_{ij}]$ is constructed, where each element $p_{ij}$ represents the conditional transition probability from state $s_i$ to state $s_j$. Formally:

$$p_{ij} = \frac{m_{ij}}{\sum_j m_{ij}}, \tag{25}$$

  where $m_{ij}$ is the number of observed transitions from $s_i$ to $s_j$.
- (3) **Estimation of the Probability Vector:** starting from an initial state $S_0$, the probability distribution vector is simulated as $\pi_t = (\pi_{1t}, \pi_{2t}, \ldots, \pi_{kt})$, where $\pi_{it}$ represents the probability that the series is in state $i$ at time $t$.
- (4) **Construction of Prediction Intervals (PIs):** based on the estimated probability distribution, the intervals are constructed by considering the central quantiles associated with the desired confidence level PINC. The prediction interval is defined as:

$$PI_t^{\text{PINC}} = [\hat{L}_t, \hat{U}_t] = \left[r_{(1-\text{PINC})/2,t},\ r_{(1+\text{PINC})/2,t}\right], \tag{26}$$

  where $\hat{L}_t$ and $\hat{U}_t$ are the estimated lower and upper bounds at time $t$, and $r_{(1\pm\text{PINC})/2,t}$ represent the respective quantiles of the estimated distribution.

The final intervals are obtained by combining the ARIMA forecasts (cline series) with the IFOMC intervals (cremaining), capturing both deterministic trend and random variations. The results showed PICP values higher than the nominal level (PINC) and high computational efficiency, being up to 3.7 times faster than QR and 7.9 times faster than LUBE, confirming the effectiveness and applicability of the method in wind power forecasting scenarios.

#### 8) A8 - WANG ET AL. [49]

In this work, interval forecasting of wind speed is implemented within a compound framework that integrates preprocessing, decomposition, and modeling stages. In the final module, responsible for interval forecasting, the aim is to capture the uncertainty of point forecasts by generating Prediction Intervals (PIs).

The approach assumes that the residuals follow a normal distribution. Thus, the interval bounds are defined from the point forecast $\hat{y}_t$ and the standard deviation $\sigma_t$ of the residuals, multiplied by a factor corresponding to the desired confidence level, as presented in the following equation:

$$PI_t = \left[\hat{y}_t - z_{\alpha/2} \cdot \sigma_t,\ \hat{y}_t + z_{\alpha/2} \cdot \sigma_t\right], \tag{27}$$

where:

- $PI_t$: prediction interval at time $t$;
- $\hat{y}_t$: point forecast generated by the compound model;
- $\sigma_t$: standard deviation of residuals at time $t$;
- $z_{\alpha/2}$: quantile of the standard normal distribution corresponding to confidence level $1 - \alpha$;
- $\alpha$: significance level (e.g., $\alpha = 0.10$ for a 90% confidence interval).

The strategy is applied to each decomposed component of the series. After the point forecast with models such as ELM, LSTM, or GRU, the standard deviation of residuals is calculated, and the PI bounds are defined using the above formula. The intervals of each component are then aggregated to form the interval of the original series.

The evaluation uses PICP (coverage) and PINAW (width). The results showed, for 90%, 95%, and 99% confidence levels: PICP = 92.31%, 96.78%, and 99.17%, with PINAW = 24.37%, 29.62%, and 40.91%, respectively. This indicates that the model generates intervals with high coverage and reduced width, suitable for short-term wind speed forecasting.

#### 9) A9 - ZHANG ET AL. [38]

The methodology uses QR to define wind speed prediction intervals. For each time $t$, the model estimates the conditional quantiles of the response variable from a linear combination of explanatory variables, according to the following equation:

$$\hat{y}_\tau(t) = \sum_{i=1}^{I} m_i x_i(t) + b, \tag{28}$$

where $x_i(t)$ represents the $i$-th explanatory variable (such as lags of the time series), $m_i$ are the coefficients to be estimated, and $b$ is the intercept. The model fitting is performed by minimizing the so-called pinball loss function, which is asymmetric and defined as:

$$E_\tau = \frac{1}{N} \sum_{t=1}^{N} \rho_\tau\left(y(t) - \hat{y}_\tau(t)\right),$$
$$\text{with} \quad \rho_\tau(u) = \begin{cases} \tau \cdot u, & \text{if } u \geq 0, \\ (\tau - 1) \cdot u, & \text{if } u < 0. \end{cases} \tag{29}$$

After estimating the lower and upper quantiles, for example with $\tau = \alpha/2$ and $\tau = 1-\alpha/2$ for a confidence level of $1-\alpha$, the prediction interval is constructed as:

$$PI_t = \left[\hat{y}_{\alpha/2}(t),\ \hat{y}_{1-\alpha/2}(t)\right]. \tag{30}$$

The interval indicates the range in which the future wind speed is expected (e.g., 90%). In addition, KDE is applied over the estimated quantiles to reconstruct the probability density function, enabling the definition of bounds and a continuous characterization of forecast uncertainty.

### 10) A10 - LI ET AL. [31]

In this hybrid model, each LSTM network generates two outputs ($u$ and $l$), interpreted as the lower and upper bounds of the prediction interval. To ensure that these bounds are correctly ordered, the following transformation is applied:

$$U = \max(u, l), \qquad L = \min(u, l). \tag{31}$$

The prediction interval is then defined as $[L, U]$, where $L$ is the lower bound and $U$ is the upper bound, thus avoiding restrictions during training and preserving the flexibility of the network.

With VMD decomposition, the series is divided into $K$ distinct modes. Each mode $i$ has its own pair of bounds $L_{i,m}$ and $U_{i,m}$, where $m$ denotes the forecast time step. The final interval is obtained through a weighted linear combination of the intervals of each mode:

$$U = \sum_{i=1}^{K} w_{i,m} \cdot U_{i,m}, \qquad L = \sum_{i=1}^{K} w_{i,m} \cdot L_{i,m}. \tag{32}$$

Here, $w_{i,m}$ represents the weight assigned to mode $i$ at time $m$, automatically adjusted through Particle Swarm Optimization. More informative modes receive higher weights to improve the final interval. These weights are optimized by a cost function that minimizes width (PINRW) and maximizes coverage (PICP), combined in the CWC metric, ensuring well-adjusted intervals and reliable forecasts for decision-making.

### 11) A11 - LI ET AL. [29]

The work proposes a hybrid model combining VMD, BPNN, and ARIMA for point and interval wind speed forecasting. The stochastic components extracted by VMD are modeled with ARIMA, followed by Monte Carlo sampling via MCMC to capture uncertainty. The prediction intervals are defined by the quantiles of the empirical distribution of the MCMC samples. The formula for the prediction interval at time $j$ is given by:

$$P_j = \left[D_j(q) + \text{pred}_j,\ U_j(1-q) + \text{pred}_j\right], \quad j = 1, 2, \ldots, N, \tag{33}$$

where:

- $\text{pred}_j$ is the point forecast of the series at time $j$;
- $D_j(q)$ represents the lower quantile (e.g., the 5th percentile) of the $M$ samples generated by MCMC simulation;
- $U_j(1-q)$ represents the upper quantile (e.g., the 95th percentile) of the same samples;
- $q$ determines the confidence level of the interval $(1-2q)$.

The proposal is to generate an adaptive interval that, by adding quantiles to the point forecast $\text{pred}_j$, remains centered on the main estimate while probabilistically capturing future variability. Performance is evaluated using Pinball Loss and Winkler Score, which consider both coverage and width, resulting in more accurate and useful forecasts for wind power systems.

### 12) A12 - SUN AND JIN [48]

Although the focus is on multi-step point forecasting, the study also presents interval forecasts (95%) as a measure of uncertainty, illustrated by shaded areas around the forecasts of the ARIMA-FNN model, suggesting symmetric intervals. However, no explicit mathematical formulation is provided, and the construction of the intervals appears to derive from model residuals, which limits reproducibility. Thus, the intervals are used only as a qualitative metric, reinforcing the robustness of the model against wind variability.

### 13) A13 - WANG ET AL. [32]

The study presents a hybrid approach combining IVMD + LSTM for point forecasting and interval forecasting based on the Lorenz Disturbance System (LDS). After the point forecast, the LDS is used to model nonlinear disturbances and, with the aid of B-splines, generate smoothed intervals.

Although the article states that LDS is used for interval construction, it does not provide formal equations describing the calculation. After the point forecast by the IVMD-LSTM model, LDS would be used to model nonlinear disturbances and form intervals around the central forecast, while B-spline functions are cited to smooth the results. However, the absence of explicit formulations limits reproducibility and prevents validation by metrics such as PICP and PINAW. In summary, the interval forecast is only described conceptually, without sufficient mathematical detail for objective verification.

### 14) A14 - ZHAO ET AL. [41]

The study proposes an ultra-short-term wind power interval forecasting approach based on CEEMDAN, Fuzzy Information Granulation, and a hybrid CNN-BiLSTM architecture. The intervals are obtained by combining the maximum, mean, and minimum sequences extracted from the fuzzy granulation of high-frequency components. For this, the triangular membership function applied to time series is used, defined as:

$$A(x) = \begin{cases} 0, & x < a \\ \dfrac{x-a}{m-a}, & a \le x \le m \\ \dfrac{b-x}{b-m}, & m < x \le b \\ 0, & x > b, \end{cases} \tag{34}$$

where $x$ represents the time variable within the window; $a$, $m$, and $b$ are, respectively, the minimum, mean, and maximum values of the series within the considered window after fuzzy granulation.

After generating the three profiles (minimum, mean, and maximum), these forecasts are combined with the non-granular residual components to form, respectively, the lower bound, the point forecast, and the upper bound of the prediction interval.

To evaluate and optimize the generated intervals, three metrics are used: PICP, PINAW, and CWC, whose formulas

are:

$$\text{PINAW} = \frac{1}{N}\sum_{i=1}^{N}\frac{U_i - L_i}{P_{t_i}} \times 100\%, \tag{35}$$

$$\text{CWC} = \begin{cases} \beta \ \text{PINAW}, & \text{if PICP} \geq \mu \\ \beta \ \text{PINAW}\big(1 + \exp\big[-\eta(\text{PICP} - \mu)\big]\big), & \text{otherwise,} \end{cases} \tag{36}$$

where $N$ is the total number of samples, $P_{t_i}$ is the actual wind power at time $t_i$, $L_i$ and $U_i$ are the lower and upper bounds of the predicted interval, $\mu$ is the nominal confidence level, $\beta$ is the weight of the interval width, and $\eta$ is a penalty parameter.

Interval generation is finalized by optimizing with CWC, balancing PINAW and PICP to meet the confidence level. This approach ensures more informative and reliable intervals, essential for decision-making in wind power generation.

#### 15) A15 - HE ET AL. [39]

The study presents a multi-step probabilistic wind speed forecasting approach based on a cooperative ensemble combining models such as SVR, GBRT, and neural networks to generate more robust forecasts. The intervals are implicitly constructed from two predicted quantiles (e.g., 5% and 95% for 90% confidence), whose outputs are combined in a weighted manner according to the performance of the base models. Although pinball loss is used, the article does not provide formal equations, describing the intervals only conceptually.

In summary, this is a probabilistic forecasting approach based on QR with ensemble, effective for representing uncertainty across multiple time horizons, but presented without detailed mathematical formulations.

#### 16) A16 - KIRTHIKA ET AL. [30]

The study proposes constructing prediction intervals using the Gauss–Newton regression, which iteratively adjusts parameters by least squares to describe the forecast trend over the lead time. From the fitted model, the confidence interval and the prediction interval (PIs) are defined based on the regression standard error and the confidence level $\tau = (1 - \alpha)$, with values of 95%, 90%, and 70% used.

After determining the wind speed bounds, the values are converted into wind power using the turbine's power curve. This is a practical and deterministic solution to quantify uncertainty with low computational cost.

#### 17) A17 - XU ET AL. [35]

The study proposes an interval forecasting model that integrates multiple upper and lower bounds generated by different GRUs, each trained independently. Because the intervals vary due to the nonlinear nature of GRUs, the results are combined via a weighted average with optimized weights.

The construction of the prediction interval is formalized by the following equations:

$$UB = \sum_{i=1}^{m}\alpha_i U_i, \tag{37}$$

$$LB = \sum_{i=1}^{m}\alpha_i L_i, \tag{38}$$

$$PI = [LB,\ UB], \tag{39}$$

where $\alpha_i$ denotes the weight assigned to the $i$-th model, and $U_i, L_i$ are the upper and lower bounds predicted by that model. The weights $\alpha_i$ are optimized by a genetic algorithm that minimizes the CWC metric, balancing reliability (coverage, PICP) and compactness (width, PINAW). This approach ensures that the produced intervals are both statistically informative and compact.

The model proposes an efficient and adaptive strategy for wind speed interval forecasting by combining time-series decomposition and GRUs. The innovation lies in the weighted integration of multiple intervals, with weights optimized via a genetic algorithm to balance accuracy and reliability.

#### 18) A18 - WANG ET AL. [51]

The study proposes a hybrid model for wind speed forecasting that combines multiscale decomposition, error correction, and GPR. In the final stage, GPR provides the interval forecast by modeling each target value as a normal random variable.

In Section 2.8 of the article, the Gaussian process is described formally. The conditional distribution of the predicted variable $y_*$, given the input $x_*$, is assumed Gaussian:

$$p(y_* \mid x_*, X, y) = \mathcal{N}(\mu_*,\ E_*), \tag{40}$$

where $\mu_*$ denotes the predictive mean and $E_*$ the associated variance. These quantities are obtained from the training set, as presented in equations (49) and (50) of the paper:

$$\mu_* = K_*^{\top}K^{-1}y \tag{41}$$

$$E_* = K(x_*, x_*) - K_*^{\top}K^{-1}K_*, \tag{42}$$

where $K$ is the covariance matrix among the training points, $K_*$ is the covariance vector between the new point $x_*$ and the training data, and $K(x_*, x_*)$ is the self-covariance at $x_*$.

Although no explicit interval formula is given, the paper infers the intervals from GPR by treating each forecast as a normal variable centered at $\mu_*$ with spread determined by $\sqrt{E_*}$, reflecting forecast uncertainty. Thus, uncertainty is incorporated directly via the estimated variance, yielding informative and reliable forecasts for wind energy applications requiring risk assessment.

#### 19) A19 - BOMMIDI ET AL. [50]

The study proposes a hybrid model for wind speed interval forecasting, combining Agile Temporal Convolutional Network and BiLSTM to capture complex temporal patterns. The lower ($L_t$) and upper ($U_t$) bounds are directly predicted by the BiLSTM for each time step $t$, based on smoothed time series. Training uses a loss function based on CWC, which balances coverage and interval width, defined as:

$$\text{Loss Function} = \arg\min_{\theta} \left( \frac{\text{CWC}(\theta)}{\text{PICP}(\theta)} \right), \quad (43)$$

where $\theta$ represents the neural network parameters. The objective is to minimize the width of the intervals (reflected in the PINRW metric) without compromising their ability to contain the actual values (PICP).

The metrics used to evaluate the intervals include PICP, while PINRW measures the mean squared width. The joint minimization of these metrics through the proposed loss function ensures that the intervals are both narrow and reliable. Thus, interval forecasting is achieved by directly estimating the bounds in a deep neural network, resulting in well-calibrated intervals applicable to wind power generation.

### 20) A20 - FU ET AL. [46]

The study proposes an interval forecasting model with Twin Support Vector Regression (TSVR) combined with the Slime Mould Algorithm (SMA), aiming to generate reliable upper and lower bounds for wind speed and address time-series uncertainty through dual regression.

The point forecast is calculated as the mean of the output functions $f(x)$ and $h(x)$ of the two TSVR regression hyperplanes:

$$F(x) = \frac{1}{2}\left[f(x) + h(x)\right]. \quad (44)$$

The prediction interval is then constructed around the point forecast $F(x)$ by applying an expansion factor based on the average distance $k$ between the regression hyperplanes:

$$k = \frac{1}{2l}\sum_{i=1}^{l} \left|f(x_i) - h(x_i)\right|. \quad (45)$$

From this value of $k$, the upper ($UB$) and lower ($LB$) bounds of the interval are determined as follows:

$$\begin{cases} UB = f(x) - a \cdot k \\ LB = h(x) + a \cdot k \end{cases}, \quad (46)$$

where $a$ is an optimization parameter tuned by the SMA, responsible for controlling the amplitude of the interval. When $a = 1$, the bounds approach the central value, resulting in a narrower interval; larger values of $a$ expand the margin to reflect higher uncertainty.

The combination of TSVR with SMA allows for the identification of optimal hyperparameters (such as $C$, $\epsilon$, and $\sigma$) for training the upper and lower models, optimizing interval forecasting accuracy. The decomposition of the series using VMD and residual modeling further strengthens interval construction, making them adaptive and well-fitted to the structure of the series and the data distribution, thereby effectively representing uncertainty in wind speed forecasting.

### 21) A21 - ABDOOS ET AL. [36]

The study proposes probabilistic prediction intervals via Monte Carlo simulation. Wind power generation data are decomposed using VMD and used to train ELM models that generate deterministic forecasts. Then, random errors, assumed to follow a normal distribution and within a predefined range (e.g., $\pm 5\%$), are added to these forecasts. The process is repeated 1000 times, producing a distribution of predicted values for each step in the forecast horizon.

From the generated sample, the lower and upper bounds of the prediction interval are calculated based on a fitted Gaussian distribution, using the mean $\bar{x}_i$ and the standard deviation $\sigma_i$ of the simulated forecasts. The interval bounds are given by:

$$L_i^{\alpha} = \bar{x}_i - z_{1-\alpha/2} \cdot \sigma_i, \quad (47)$$

$$U_i^{\alpha} = \bar{x}_i + z_{1-\alpha/2} \cdot \sigma_i, \quad (48)$$

where $z_{1-\alpha/2}$ is the critical value of the standard normal distribution corresponding to the desired confidence level ($\alpha$).

Subsequently, the quality of the prediction intervals is evaluated using different metrics, including PICP, PINAW, and ACE (Average Coverage Error), which represents the difference between the nominal confidence level (PINC) and the PICP. Interval Sharpness and CRPS are also highlighted, as they combine aspects of reliability and accuracy of the intervals.

This approach captures the uncertainty of wind power forecasting, providing reliable intervals that support decision-making in energy systems.

### 22) A22 - ZHANG ET AL. [33]

The prediction interval is constructed with Smoothed Kernel Density Estimation (SKDE) applied to model errors, representing uncertainty nonparametrically. After training with historical data, errors are extracted and, to address sample limitations, augmented via SMOTE, which generates synthetic samples by interpolating between nearest neighbors. The new errors are obtained as:

$$e_{\text{ijnew}} = e_i + r_j(e_j - e_i), \quad (49)$$

where $e_i$ and $e_j$ are real error samples, and $r_j \in [0, 1]$ is a random number.

With the augmented error set, SKDE is applied to estimate the Probability Density Function (PDF) of the errors. The estimate is given by:

$$\hat{f}_m(x) = \frac{1}{nm}\sum_{i=1}^{n} K\left(\frac{x - x_i}{m}\right), \quad (50)$$

where:

- $n$ is the number of error samples (including synthetic ones),
- $m$ is the smoothing parameter (bandwidth),

- $K(\cdot)$ is the kernel function, assumed Gaussian in this work:

$$K(u) = \frac{1}{\sqrt{2\pi}} \exp\left(-\frac{u^2}{2}\right). \tag{51}$$

From the estimated PDF, the interval bounds are obtained by integrating the density until the area corresponds to $1 - \alpha$, forming symmetric intervals around the forecast. This approach flexibly and nonparametrically captures the distribution of errors, proving effective in noisy and nonlinear data such as wind forecasting.

#### 23) A23 - WANG ET AL. [40]

The prediction intervals are constructed by applying KDE to the point forecast errors obtained with a GRU network. KDE, a nonparametric density estimation method, models the PDF of errors simply and without assuming distributions. The estimate is given by:

$$\hat{f}(e) = \frac{1}{OU} \sum_{o=1}^{O} K\left(\frac{e - e_o}{U}\right), \tag{52}$$

where:

- $\hat{f}(e)$ is the estimated probability density of error $e$;
- $O$ is the total number of error samples;
- $U$ is the bandwidth, determined via grid search;
- $e_o$ represents the $o$-th observed error value;
- $K(\cdot)$ is the kernel function; in this study, a box kernel was used.

After obtaining the PDF, the CDF is derived by integrating the estimated density. From the CDF, the quantiles corresponding to a desired confidence level $1 - \alpha$ are determined, thus defining the lower and upper bounds of the prediction interval. The construction of the interval is formalized as:

$$P(G_{\text{low}} < G < G_{\text{up}}) = 1 - \alpha, \tag{53}$$

where:

- $G$ is the offshore wind speed forecast value;
- $G_{\text{low}}$ and $G_{\text{up}}$ are the lower and upper bounds of the prediction interval;
- $1 - \alpha$ is the desired confidence level (e.g., 85% or 95%).

This procedure enables generating reliable prediction intervals adjustable to different confidence levels, without assuming a specific functional form for the error distribution. Applying KDE to residuals provides a more accurate characterization of uncertainty, reflecting the actual variations present in the input data.

#### 24) A24 - LI ET AL. [65]

The study quantifies the uncertainty of point forecasts through prediction intervals (PIs), assuming normally distributed errors. The wind series is decomposed via CEEMDAN into Intrinsic Mode Functions (IMFs), each forecasted by a CNN-LSTM network. The final series is reconstructed as a linear combination of the forecasted IMFs, according to the regression model:

$$\hat{y} = X\hat{\beta}, \tag{54}$$

where $\hat{y}$ represents the vector of reconstructed forecasts, $X$ is the matrix composed of forecasted IMFs, and $\hat{\beta}$ is the vector of estimated coefficients.

Assuming Gaussian residuals, the prediction interval with confidence level $100(1 - \alpha)\%$ for each forecasted point is given by:

$$\begin{cases} L_i = \hat{y}_i - B \\ U_i = \hat{y}_i + B \\ B = z_{\alpha/2} \cdot \hat{\sigma} \cdot \sqrt{1 + x_i^\top (X^\top X)^{-1} x_i}, \end{cases} \tag{55}$$

where:

- $L_i$ and $U_i$ are the lower and upper bounds of the prediction interval for the $i$-th observation;
- $z_{\alpha/2}$ is the quantile of the standard normal distribution corresponding to significance level $\alpha$ (e.g., $z_{0.025} \approx 1.96$ for 95% confidence);
- $\hat{\sigma}$ is the estimated standard deviation of the residuals;
- $x_i$ is the row vector corresponding to the $i$-th row of $X$.

The standard deviation of the residuals is estimated as:

$$\hat{\sigma} = \sqrt{\frac{1}{n - K - 1} \sum_{i=1}^{n} (y_i - \hat{y}_i)^2}, \tag{56}$$

where $n$ is the total number of samples and $K$ is the number of parameters (IMFs considered).

The results show that CEEMDAN–CNN–LSTM improves accuracy and generates narrower intervals: an average reduction of 7% compared to CEEMDAN–LSTM and up to 20% compared to other models, confirming its superiority in quantifying wind speed uncertainty.

#### 25) A25 - ZHAO ET AL. [37]

The study proposes a hybrid strategy that combines EWT with correntropy-optimized algorithms (RCVR, LSSVM, and ELM) to improve short-term wind speed forecasting. Although it employs probabilistic interval forecasting evaluated with PICP and PINAW, it does not describe how the intervals were constructed: there is no mention of methods such as bootstrap, conformal prediction, or Bayesian approaches, nor of assumptions about residual distributions. The absence of equations and methodological details limits reproducibility and comparability with other approaches.

## VI. FUTURE RECOMMENDATIONS

Future research should advance the development of hybrid interval forecasting models that integrate deep learning with probabilistic approaches. These models must also be evaluated in terms of computational cost, sensitivity to parameter changes, and real-world applicability to ensure their practical viability.

There is a notable concentration of studies focused on short-term forecasting. Expanding research into medium- and long-term horizons—considering seasonal patterns and extended historical data—would support more strategic planning in wind energy systems.

TABLE 3. Acronyms list.

| Description | Acronym |
|---|---|
| Auto-Regressive Integrated Moving Average | ARIMA |
| CKDE-NRC (Conditional Kernel Density Estimation with bandwidth determined by the normal reference criterion) | CKDE-NRC |
| Complete Ensemble Empirical Mode Decomposition with Adaptive Noise | CEEMDAN |
| Convolutional Neural Network-Bidirectional Long Short-Term Memory | CNN-BiLSTM |
| Continuous Ranked Probability Score | CRPS |
| Coverage Width Criterion | CWC |
| Empirical Mode Decomposition | EMD |
| Ensemble Empirical Mode Decomposition with Adaptive Noise | ICEEMDAN |
| Extreme Learning Machine | ELM |
| Gated Recurrent Unit | GRU |
| Gaussian Process Regression | GPR |
| Graph Convolutional Networks | GCN |
| Improved First-Order Markov Chain | IFOMC |
| Interval Forecast Coverage Probability | IFCP |
| Interval Forecast Normalized Average Width | IFNAW |
| Intrinsic Mode Functions | IMFs |
| Kernel Density Estimator | KDE |
| Latent Dirichlet Allocation | LDA |
| Long Short-Term Memory | LSTM |
| Lorenz Disturbance System | LDS |
| Machine Learning | ML |
| Mean Absolute Error | MAE |
| Mean Absolute Percentage Error | MAPE |
| Mixture Density Neural Network | MDNN |
| Monte Carlo Markov Chain | MCMC |
| Normalized Mean Prediction Interval Width | NMPIW |
| Prediction Interval Coverage Probability | PICP |
| Prediction Interval Normalized Average Width | PINAW |
| Prediction Interval Normalized Root-mean-square Width | PINRW |
| Probability Density Function | PDF |
| Quantile Linear Regression | QLR |
| Quantile Regression | QR |
| Quantile Regression Forest | QRF |
| Refined Composite Multiscale Fuzzy Entropy | RCMFE |
| Recurrent Extreme Learning Machines | RELM |
| Root Mean Square Error | RMSE |
| Self-Organizing Map | SOM |
| Singular Spectrum Analysis | SSA |
| Slime Mould Algorithm | SMA |
| Smoothed Kernel Density Estimation | SKDE |
| Static Line Rating | SLR |
| Twin Support Vector Regression | TSVR |
| Variational Mode Decomposition | VMD |
| Wavelet Soft Threshold | WST |

Another research priority is the validation of models using real-time and multisource data, such as IoT sensors and satellite-based meteorological information. This would enhance the robustness and generalization of forecasting models beyond specific wind farms.

Additionally, greater consistency is needed in the use of performance metrics for interval forecasting. While accuracy remains important, future work should emphasize measures that capture uncertainty, such as PICP, CWC, and CRPS. Standardizing evaluation practices across studies would improve comparability and reduce misinterpretations caused by dataset or metric variation.

In summary, the future of wind energy forecasting depends on models that not only improve predictive performance but also offer interpretable, reliable, and actionable intervals under realistic operating conditions.

## VII. CONCLUSION

This study presented a comprehensive review of interval forecasting models applied to wind power generation, emphasizing preprocessing techniques, forecasting methods, and performance metrics.

Among preprocessing methods, VMD was the most widely used, followed by CEEMDAN and EMD, reflecting their effectiveness in handling non-stationary and noisy time series. In forecasting techniques, LSTM was the most adopted model, with traditional methods like ARIMA and KDE also remaining relevant, especially in studies evaluated using point-based metrics such as RMSE and MAE.

Although interval-based performance measures such as PICP, CWC, and CRPS were less frequently used, their presence highlights a growing concern with modeling forecast uncertainty. These metrics were mainly associated with kernel-based methods and deep learning models that generate probabilistic outputs.

The review also revealed challenges, particularly the lack of standardization in evaluation practices, which limits comparability across studies. Despite the technical feasibility of benchmarking multiple models on the same datasets, few works follow this practice. Moreover, while interval metrics provide richer insights, their computational complexity may hinder broader adoption.

Future research should focus on developing models capable of producing reliable and well-calibrated prediction intervals, reducing uncertainty while maintaining precision. Standardizing the use of evaluation metrics will be essential for advancing reproducibility and improving the quality of comparative analyses. As wind energy becomes increasingly relevant in modern power systems, robust interval forecasting methods will play a key role in supporting more efficient and resilient energy planning.

## ACKNOWLEDGMENT

The authors thank CAPES for enabling open access through its transformative agreement with IEEE. Coordination for the Improvement of Higher Education Personnel – CAPES (ROR identifier: 00x0ma614).